\documentclass{article}
\usepackage{iclr2027_conference,times}
\iclrfinalcopy  

\usepackage[utf8]{inputenc}
\usepackage[T1]{fontenc}
\usepackage{hyperref}
\usepackage{url}
\usepackage{booktabs}
\usepackage{amsfonts}
\usepackage{amsmath}
\usepackage{amssymb}
\usepackage{microtype}
\usepackage{xcolor}
\usepackage{graphicx}
\usepackage{multirow}
\usepackage{subcaption}

\newif\ifdraftmode
\draftmodetrue

\newcommand{\xhdr}[1]{\vspace{0em}\noindent{{\bf #1.}}}

\newcommand{\lockedLEDOVrecovVal}{55.6\,\%}    
\newcommand{\lockedLEDOVrecovTest}{56.4\,\%}   
\newcommand{\lockedSalTemptoRecovVal}{15.8\,\%}
\newcommand{\lockedSalTemptoRecovTest}{10.2\,\%}
\newcommand{\lockedLEDOVbestIGE}{75.4\,\%}     
\newcommand{\lockedLEDOVbestIGEtest}{75.2\,\%} 
\newcommand{\lockedSalTemptoBestIGEval}{55.6\,\%}
\newcommand{\lockedSalTemptoBestIGEtest}{61.4\,\%}
\newcommand{\lockedLEDOVrecovPooled}{56\,\%}      
\newcommand{\lockedSalTemptoRecovPooled}{13\,\%}  
\newcommand{\lockedLEDOVbestIGEpooled}{75\,\%}    
\newcommand{\lockedSalTemptoBestIGEpooled}{58.5\,\%} 
\newcommand{\lockedRho}{0.74}                      
\newcommand{\lockedRhoExcluded}{0.83}              
\newcommand{\lockedLEDOVgapPP}{19}                 
\newcommand{\lockedSalTemptoGapPP}{45}             

\newcommand{\lockedSalFoMresid}{0.77\,bit}        
\newcommand{\lockedSalFoMresidpct}{41.5\,\%}      
\newcommand{\lockedFrameMeanLEDOV}{+0.56\,bit}    
\newcommand{\lockedFrameMeanSalTempto}{+0.77\,bit}
\newcommand{\lockedFrameTailLEDOV}{+1.67\,bit}    
\newcommand{\lockedFrameTailSalTempto}{+2.4\,bit}
\newcommand{\lockedFrameSkewLEDOV}{+4.72}
\newcommand{\lockedFrameSkewSalTempto}{+2.40}
\newcommand{\sigmaCBLEDOV}{4.44\,\%}
\newcommand{\sigmaCBSalTempto}{3.15\,\%}

\newcommand{\Nvideos}{224}
\newcommand{\Nsubjects}{16}
\newcommand{\Ntrain}{204}
\newcommand{\Nval}{10}
\newcommand{\Ntest}{10}

\newcommand{\Nmodels}{13}

\newcommand{\datasetname}{SalTempto}

\title{Moving Forward with Video Saliency: A New Dataset and Benchmark where Motion Matters}

\author{\parbox[t]{\dimexpr\textwidth-2\tabcolsep\relax}{\centering\normalfont
  \textbf{Susmit~Agrawal}$^{1,2}$ \quad \textbf{Rebecca~Wanner}$^{3}$ \quad
  \textbf{Juliane~Verwiebe}$^{1,2}$ \quad \textbf{Matthias~Tangemann}$^{4,5}$ \\[0.5em]
  \textbf{Matthias~Bethge}$^{1}$ \qquad \textbf{Matthias~K\"ummerer}$^{1}$ \\[0.8em]
  $^{1}$T\"ubingen AI Center, University of T\"ubingen \quad $^{2}$IMPRS-IS \\
  $^{3}$Optocycle \quad $^{4}$University of Toronto \quad $^{5}$Vector Institute}}

\begin{document}

\maketitle
\lhead{Preprint. Under review.}
\begin{abstract}
Video saliency prediction attempts to capture more natural human visual behaviour than static image saliency, and is inherently harder to model due to the additional temporal dimension. Video saliency benchmarks rest on an implicit premise that predicting gaze on video requires utilizing temporal activity distributed across frames. Prior work has challenged this premise, showing that a deliberately static baseline recovers a significant fraction of the explainable gaze information on LEDOV, a popular video saliency dataset, without integrating temporal context, and that video saliency models fail in the same places as this static baseline. We first verify that this diagnosis still stands: under a more capable gold standard than the original analysis used, and an updated panel of recent architectures, the strongest temporal architecture in the panel still does not seem to substantially improve over a fine-tuned static baseline. However, it remains unclear whether the marginal gain reflects limitations of current temporal architectures or a lack of genuinely temporal patterns in the benchmark itself. We introduce SalTempto, a new video saliency benchmark with greater dynamism: 224 one-minute clips of highly dynamic content, sourced from the HACS-Segments action dataset so that each clip contains an event together with its lead-up and aftermath, with gaze recordings from up to 16 subjects and a training split for adapting pretrained models. On SalTempto, the same fine-tuned static baseline recovers only about 13\% of the headroom above the centerbias, against more than half on LEDOV. The strongest fine-tuned temporal architecture shows a substantial gain in performance over the static baseline, indicating that it does capture meaningfully more temporal information, which LEDOV fails to measure. Yet, even this SoTA model still leaves nearly half of SalTempto's headroom unexplained, indicating that there is still a lot of room for improvement in the field of video saliency. In addition, manual examination of SalTempto recordings allows us to describe three human visual tendencies that models tend to miss: \emph{object permanence despite occlusion}, \emph{scene inertia}, and \emph{anticipatory saccades}. SalTempto releases with raw visual recordings and training and evaluation videos. Link to anonymous repository: \url{https://huggingface.co/datasets/bethgelab/video_saliency}.
\end{abstract}

\section{Introduction}
\label{sec:intro}

Video saliency models predict where people tend to look when viewing a video. Their predictions serve as attention priors in content-aware video compression, foveated rendering, and evaluation of visual interfaces, and they are the primary computational account of how human attention responds to events as they unfold. What separates the video problem from image saliency is this temporal component: a video model should use viewing history together with scene dynamics to predict where human attention will flow, not only what draws attention in the current frame being observed. Video saliency models are accordingly built on temporal architectures such as recurrent backbones, 3D convolutions, and transformers. They rely on the implicit premise that predicting visual saliency in video requires reasoning across frames. \citet{tangemann_measuring_2020} tested this premise on LEDOV \citep{jiang_predicting_2018}, reporting that a static baseline recovers at least 75\,\% of the net explainable gaze information (based on a gold standard model setting an upper bound estimate on performance) without integrating temporal information. Their probe outperformed the state-of-the-art temporal models of the time on standard saliency map evaluation metrics; visual inspection of its failure cases identified clear temporal effects on gaze that are present in LEDOV but rare and not captured by the evaluated panel of models. The implication is uncomfortable: progress on LEDOV cannot reliably be read as progress on temporal modelling, with evaluations on other video saliency datasets presenting their own problems as discussed in later sections. The benchmark needed to test the temporal capabilities of video saliency models requires a stand-alone dataset with greater dynamism, on which static features cannot already do most of the work. 

Here, we verify that \citeauthor{tangemann_measuring_2020}'s diagnosis still holds despite progress in video saliency models that incorporate temporal aspects. Against a more capable gold standard than the original analysis used (\S\ref{sec:methods}), and an updated panel of \Nmodels{} architectures now including post-2020 video saliency models, the static baseline still recovers \lockedLEDOVrecovPooled{} of the headroom above centerbias on LEDOV. The strongest current temporal model reaches \lockedLEDOVbestIGEpooled{}, showing only a \lockedLEDOVgapPP{}-point improvement over the static baseline. It remains unclear whether this marginal gap reflects limitations of current temporal architectures or a lack of genuinely temporal patterns in LEDOV itself. We perform additional analysis on LEDOV's fixation distributions, and find that gaze targets cluster spatially even across time within LEDOV videos. Consequently, one frame's gold-standard density predicts nearly 30\% of the net available gaze information of another arbitrary frame in a video (App.~\ref{app:dynamism:crossframerecov}). In other words, models usually do not face the challenge of tracking salient regions that drift spatially over time.

To overcome these problems, we introduce \datasetname{}: \Nvideos{} one-minute clips of highly dynamic content, sourced from HACS-Segments \citep{zhao_hacs_2019} so that each clip contains an annotated event together with its lead-up and aftermath, with gaze recordings from up to \Nsubjects{} free-viewing subjects. The resulting foreground dynamics yield wider within-video gaze dispersion than LEDOV. On \datasetname{}, gaze targets shift between frames to a degree that models must actually attend to salient content over time to meaningfully predict gaze targets (App.~\ref{app:dynamism}). A fine-tuned static baseline accordingly recovers a much smaller fraction of the available gaze information than on LEDOV, while the strongest temporal architecture shows a substantial gain over the static baseline. This indicates that current architectures measurably capture temporal structure when given a benchmark that surfaces it. Yet, the same architecture leaves $\approx$\lockedSalFoMresidpct{} of \datasetname{}'s headroom unexplained (\S\ref{sec:results:residual}), indicating that the video saliency problem is far from solved. In addition, the length of frames, diversity of content, and dense gaze recordings allow us to identify three additional viewer behaviours on video data that models tend to miss: \emph{object permanence despite occlusion}, \emph{scene inertia}, and \emph{anticipatory saccades} (\S\ref{sec:results:failures}).


\section{Related work and existing datasets}
\label{sec:related}

Among recent video saliency benchmarks like DHF1K \citep[1000 videos, 17 subjects]{wang_revisiting_2018}, DIEM \citep{mital_clustering_2011}, Hollywood-2 and UCF-Sports \citep{mathe_actions_2015}, the LEDOV dataset \citep[538 videos, 32 subjects]{jiang_predicting_2018} has become a preferred choice for fine-tuning and evaluating new architectures. Each of the other video saliency benchmarks falls short on at least one criterion essential to the diagnostic. \textbf{DHF1K} data is known to have artefacts in the published fixation masks \citep{tangemann_measuring_2020}. Additionally, DHF1K's released gaze maps do not preserve per-subject identification, preventing the leave-one-subject-out gold-standard computation that the diagnostic relies on (\S\ref{sec:methods}). \textbf{Hollywood-2} and \textbf{UCF-Sports} \citep{mathe_actions_2015} record gaze under task-aware observation rather than free-viewing, as subjects were performing action recognition during data collection. They also confine their stimulus distributions to a single content domain (movie scenes for Hollywood-2; sporting scenarios for UCF-Sports), limiting their reach as general-purpose video saliency benchmarks. \textbf{DIEM} (84 videos) was recorded with audio playback and heterogeneous recording setups; we report a zero-shot comparison on it in App.~\ref{app:diem} for completeness despite the confounders. LEDOV is the original target for prior work, the preferred choice for fine-tuning new video saliency architectures, and a free-viewing video saliency benchmark.

\citet{tangemann_measuring_2020} introduced the static-recoverability diagnostic together with DeepGaze MR (DG-MR), a static baseline built for the measurement, and applied both to LEDOV with the result summarised in \S\ref{sec:intro}. To direct subsequent development, they assembled a meta-dataset of temporally demanding examples drawn from existing fixations on existing clips; a stand-alone benchmark with greater dynamism, on which a static baseline cannot already do most of the work, was not part of that contribution and is what \datasetname{} provides.

\section{Methods}
\label{sec:methods}

\begin{figure}[t]
\centering
\includegraphics[width=\linewidth]{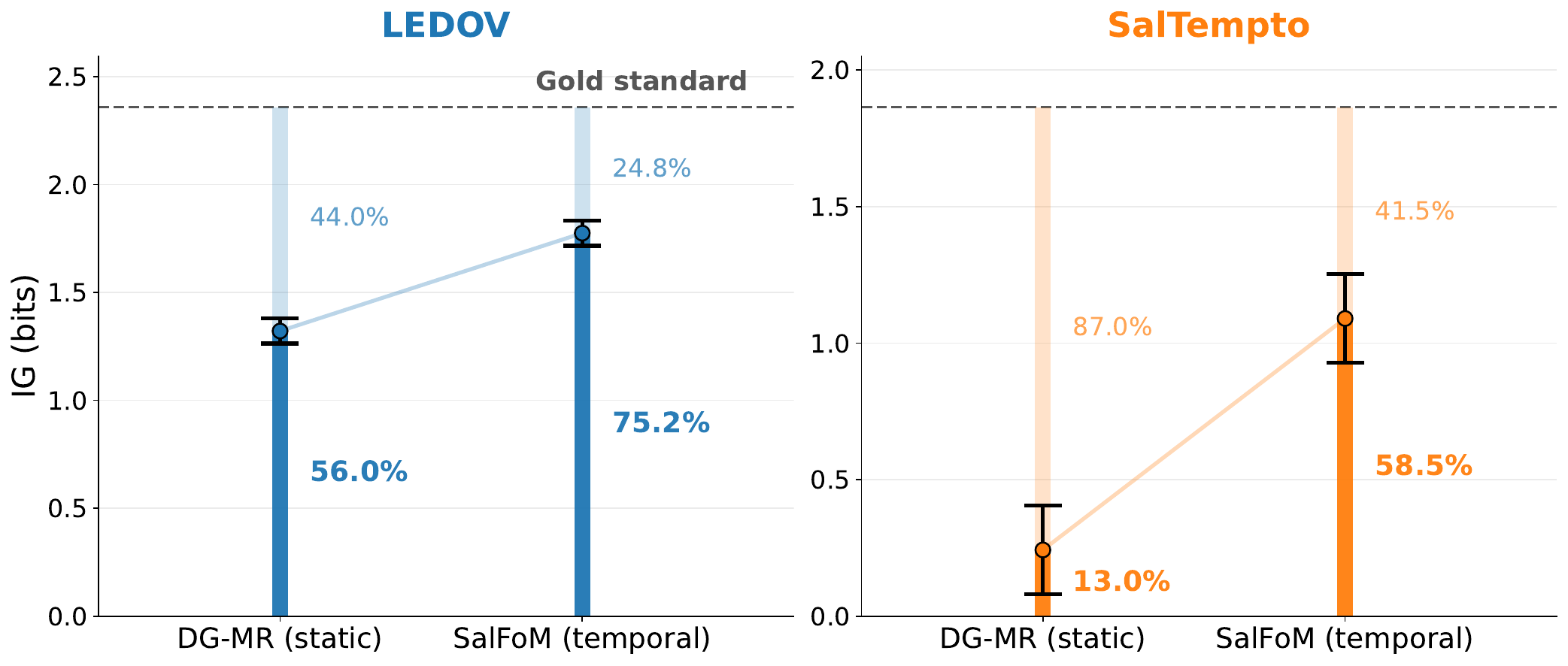}
\caption{Static-vs-temporal performance and remaining headroom on LEDOV (left) and \datasetname{} (right). Each panel shows the dataset's IG over centerbias LL on the $y-$axis and the gold-standard IG ceiling (dashed) at the top. Each model's bar is split into a closed segment (solid; percent of headroom recovered) and a remaining segment (faded; percent unexplained). Both models are scored on the same videos, so error bars on the model markers are 95\,\% intervals for the paired comparison between them, obtained by removing each video's mean over the two models \citep{cousineau_confidence_2005,morey_confidence_2008}; intervals for absolute IG and for the gold ceiling are in Table~\ref{tab:jackknife}. The static-vs-temporal climb (faded line connecting DG-MR to SalFoM) is steeper on \datasetname{} than on LEDOV, indicating that SalFoM extracts a lot more information than the static baseline on SalTempto than on LEDOV.}
\label{fig:finetuned_ig_with_gaps}
\end{figure}

\subsection{\datasetname{} Dataset Creation}
\label{sec:dataset}

\xhdr{Source dataset and stimulus selection} An important prerequisite for capturing temporal effects is the use of videos that
are long enough to contain temporal variations. We obtain this structure from HACS-Segments \citep{zhao_hacs_2019}, a video dataset spanning 200 categories whose annotations mark the start and end of recognisable actions within each clip. Actions, including their beginning and ending, inherently
involve temporal dynamics, leading to highly dynamic and varied temporal effects being present in the final
dataset. Videos from this dataset also contain other dynamic events like camera motion or scene cuts, which are viewed by observers and contribute to gaze dynamics. The broad range of categories further ensures greater variance in the presented stimuli. From the HACS pool we filter by resolution ($\geq 1280 \times 720$) and sample \Nvideos{} clips satisfying action-segment positioning constraints: each clip contains an annotated action segment beginning at $\geq$15\,s, with midpoint $\geq$30\,s, and ending before clip end. This construction yields measurably greater within-video gaze dispersion than LEDOV (App.~\ref{app:dynamism}).
\label{sec:dataset:source}


\xhdr{Fixation collection}
\Nsubjects{} participants (9 female, 7 male; mean age 28.6) with normal
or corrected-to-normal vision contributed gaze recordings with informed consent. Each training video received gaze from 3 subjects, while each validation and test video was viewed by all \Nsubjects{} subjects. This reflects different tradeoffs between video diversity and density estimation quality: for training models, we benefit from having diverse videos at the cost of subject diversity per video (which is unproblematic for point objectives like fixation log-likelihood). Additionally, per-subject viewing tendencies do not contribute significant information \citep{agrawal2026deepgaze3.5} especially under unknown viewer identity, therefore subject biases are negligible. On the validation and test splits, we need enough samples for estimating gold standard densities for reliable analyses and case studies, hence we trade-off video quantity for more gaze samples under constant budget, as is common in such studies \citep{schwetlick2024potsdam}. Subjects viewed $1920 \times 1080$ stimuli on a VIEWPixx 3D monitor refreshed at 120\,Hz, head-stabilised by a 50\,cm chin rest, while their gaze was sampled at 500\,Hz by an EyeLink 1000 calibrated with a 5-point procedure and drift-checked between blocks (off-screen samples excluded). There was no audio present in the stimuli. When starting the experiment, a 5-point calibration was performed, followed up by a
validation. Every participant was instructed to view 50 videos, without giving any task. Each stimulus
was starting with a fixation cross in the center. Between two stimuli, the participant had the choice to take a break as long as they want and before continuing
a drift check was performed by displaying a fixation cross at the center. If the measured gaze position
was off by >2°, the calibration was repeated to ensure accurate gaze positions.
The stimuli videos for each subject were separated into 20 validation videos and 30 training videos.
It was randomized if the trial starts with a training or validation
video. Additionally, the order of the videos was randomized to minimize potential order effects like
fatigue, contextual biases and stimulus-specific biases.

Per-frame gaze positions used in the analyses of \S\ref{sec:results} are obtained by binning the per-subject gaze stream into video frames, preserving subject identity.  The leave-one-subject-out gold-standard computation in \S\ref{sec:methods:goldstandard} also operates on the per-subject binned positions. \datasetname{} releases the underlying raw 500\,Hz trajectories alongside the binned positions, so downstream consumers can recover the saccade and smooth-pursuit dynamics that frame-averaging or fixation-stabilisation would discard.
\label{sec:dataset:eye}

\xhdr{Splits and release}
\datasetname{} is split into \Ntrain{} training videos, \Nval{} validation videos, and \Ntest{} test videos. In video-hours this is 3.4\,h of training video and 0.17\,h each of validation and test video, against 1.52\,h and 0.13\,h for LEDOV; in observer-hours, $\approx$8.6\,h of training gaze and $\approx$2.7\,h each for validation and test. Training and validation videos are released publicly with eye-tracking data for both eyes, and additional data such as pupil sizes and sample timestamps at every recording. The test split videos are publicly available, with held-out recordings providing a reference point that guards against benchmark overfitting and lets the field track whether the static-recoverable gap closes as models evolve. On acceptance, \datasetname{} will be released as a hosted video saliency benchmark.
\label{sec:dataset:splits}

\subsection{The SalTempto Benchmark}

\xhdr{Models and fine-tuning protocol}
\label{sec:methods:framework}
We benchmark \Nmodels{} architectures: DeepGaze MR (DG-MR), the static baseline introduced by \citet{tangemann_measuring_2020}, and twelve temporal models spanning from older architectures to the recent state-of-the-art models: SalFoM \citep{moradi_salfom_2024}, TMFI-Net \citep{zhou_tmfinet_2023}, ViNet-\{S, A, E\} \citep{jain_vinet_2021}, TASED-Net \citep{min_tasednet_2019}, TinyHD-\{single, multi\} \citep{hu_tinyhd_2023}, UniformerSal-\{Spatial, SpatioTemporal\} \citep{li_uniformer_2022}, and UNISAL-\{Spatial, SpatioTemporal\} \citep{droste_unisal_2020}. The released checkpoints for DG-MR and UniformerSal were trained on LEDOV by their original authors; the remaining ten models' released checkpoints were trained primarily on DHF1K.

All models are evaluated zero-shot from their released checkpoints. In addition, we fine-tune both DG-MR (static baseline) and SalFoM on each dataset's training split and evaluate them as well. We select SalFoM as the temporal architecture for fine-tuning because it is the strongest model in the panel. We did not run additional fine-tuning of DG-MR on LEDOV because the released checkpoint was already trained on LEDOV. Fine-tuning ran for roughly 10\,h (DG-MR) and 12\,h (SalFoM) on a single NVIDIA A100; all other evaluations are inference only. As a within-architecture control, we also evaluate the fine-tuned SalFoM with its temporal context removed, repeating the target frame across the model's full context window so that the input carries no motion (App.~\ref{app:static-salfom}).

\xhdr{Centerbias}
Humans have a tendency to fixate at the center of a scene. The data collection pipeline further exacerbates this tendency by starting each recording session with a central fixation. This centerbias serves as a fundamental prior in studies of visual attention, as it is easy to model and accounts for both data collection biases and natural viewing behaviour. Furthermore,  benchmarking against the centerbias ensures that models are evaluated based
on the ability to predict meaningful information, rather than exploiting trivial spatial biases \citep{kummerer_deepgaze_2016,kummerer_predicting_2023,agrawal2026raisingceilingbetterempirical}. Each dataset's centerbias is a Gaussian KDE on training fixations, with bandwidth $\sigma$ chosen by grid search on held-out log-likelihood: $\sigma = \sigmaCBSalTempto{}$ on \datasetname{}, $\sigmaCBLEDOV{}$ on LEDOV. The different optima reflect each dataset's fixation spread, not a methodological asymmetry.

\xhdr{Gold standard} 
\label{sec:methods:goldstandard}
Following the recently proposed method for estimating inter-observer consistency by \citet{agrawal2026raisingceilingbetterempirical}, our 
 per-frame gold standard is a 4-component mixture consisting of leave-one-subject-out Abramson KDE, a uniform prior, the dataset's centerbias, and the fine-tuned SalFoM model predictions. The mixture weights are chosen per frame by the held-out-subject fixation likelihood. The gold standard is an estimator of maximum achievable performance. While the KDE captures spatial consistency across observers within a frame, the SalFoM model is used to account for more semantic and other consistencies (such as temporal aspects) across frames and videos. As noted by \citet{agrawal2026raisingceilingbetterempirical}, without the deep learning component in the mixture, the upper bound estimate achieves a LL of approximately 2 bits/fix on LEDOV; SalFoM achieves approximately 2.9 bits/fix \emph{in zero-shot evaluation}, and exceeds 3.0 bits/fix post fine-tuning, showing that accounting for semantic, cross-frame and video consistency is crucial for making the ``upper bound'' trustworthy.

\xhdr{Evaluation metrics} Following \citet{kummerer_information-theoretic_2015,kummerer_predicting_2023}, we base the bulk of our analysis and evaluations on information-theoretic metrics. Specifically, we use the Log-Likelihood (LL) predicted by a model at a ground-truth gaze point as a measure of how well it predicts the region to be salient to humans. This also allows a direct comparison against the centerbias prior: the Information Gain (IG) a model captures over the centerbias can be expressed naturally as $IG = LL(model) - LL(centerbias)$.
The gap between the centerbias (trivially achievable performance) and the gold standard (upper performance bound) a model captures is summarized via Information Gain Explained (IGE; defined in App.~\ref{app:results-tables}). The gold standard enters only the denominator of IGE, which is shared by every model on a dataset; LL, IG, and the resulting model rankings do not depend on it. We additionally report standard saliency-evaluation metrics in the same appendix: KL divergence (KLD) against a Gaussian-blurred fixation density map, Pearson correlation (CC), histogram similarity (SIM), area under the ROC curve (AUC), and normalized scanpath saliency (NSS); see \citet{bylinskii_what_2019} for definitions and a comparative analysis.

\section{Dataset Analysis and Results}
\label{sec:results}

\xhdr{The static-vs-temporal gap is more than twice as large on \datasetname{} as on LEDOV}
\label{sec:results:gap}
We find that the strongest temporal model (SalFoM-Tuned) exceeds the static baseline (DG-MR-Tuned) by \lockedSalTemptoGapPP{}\, percentage points on \datasetname{} (aggregated across both validation and test splits; visualized in Figure~\ref{fig:finetuned_ig_with_gaps}, right; per-split values in Table~\ref{tab:matched_protocol}). On LEDOV, the same comparison gives a \lockedLEDOVgapPP{}-pp gap (Figure~\ref{fig:finetuned_ig_with_gaps}, left). The static-vs-temporal separation is therefore more than twice as large on \datasetname{} as on LEDOV: the temporal architecture captures information not available to a static baseline, but the LEDOV benchmark does not surface this difference as clearly. The gap is not a matter of stronger static features alone: with its temporal context removed (\S\ref{sec:methods:framework}), SalFoM recovers 33\,\% of \datasetname{}'s headroom against \lockedSalTemptoBestIGEpooled{} with context, so 44\,\% of its information gain over the centerbias depends on temporal context; on LEDOV the same ablation leaves 64\,\% recovered, with only 15\,\% of its gain depending on temporal context (App.~\ref{app:static-salfom}).

\xhdr{Fine-tuning adapts pretrained models rather than adding capability}
Fine-tuning on the \datasetname{} training split raises SalFoM by about four IGE points on each split and lifts DG-MR from at or below the centerbias to \lockedSalTemptoRecovPooled{} (App.~\ref{app:results-tables}). Both gains are consistent with adaptation to a new recording setup rather than with newly acquired capability, so the training split serves to adapt pretrained models, and the remaining headroom is not explained by a shortage of training fixations.

\xhdr{The strongest temporal architecture leaves nearly half of \datasetname{}'s headroom unexplained}
Even after fine-tuning, SalFoM's residual to gold on \datasetname{} remains \lockedSalFoMresid{} ($\approx$\lockedSalFoMresidpct{} of the headroom above centerbias; Figure~\ref{fig:finetuned_ig_with_gaps} (right)). This is in contrast to LEDOV, where less than 25\% is left unexplained by SalFoM.

\xhdr{On LEDOV, DG-MR explains \lockedLEDOVrecovPooled{} of the explainable information gain estimated by the gold standard} The gap compared to \citet{tangemann_measuring_2020} arises from the gold standard, not the metric: where their gold standard is a leave-one-subject-out KDE mixed with a uniform distribution, ours is a 4-component mixture that includes the dataset's centerbias and a fine-tuned SalFoM component (\S\ref{sec:methods}).

\begin{figure}[ht]
\centering
\begin{subfigure}[t]{0.62\linewidth}
  \centering
  \includegraphics[width=\linewidth]{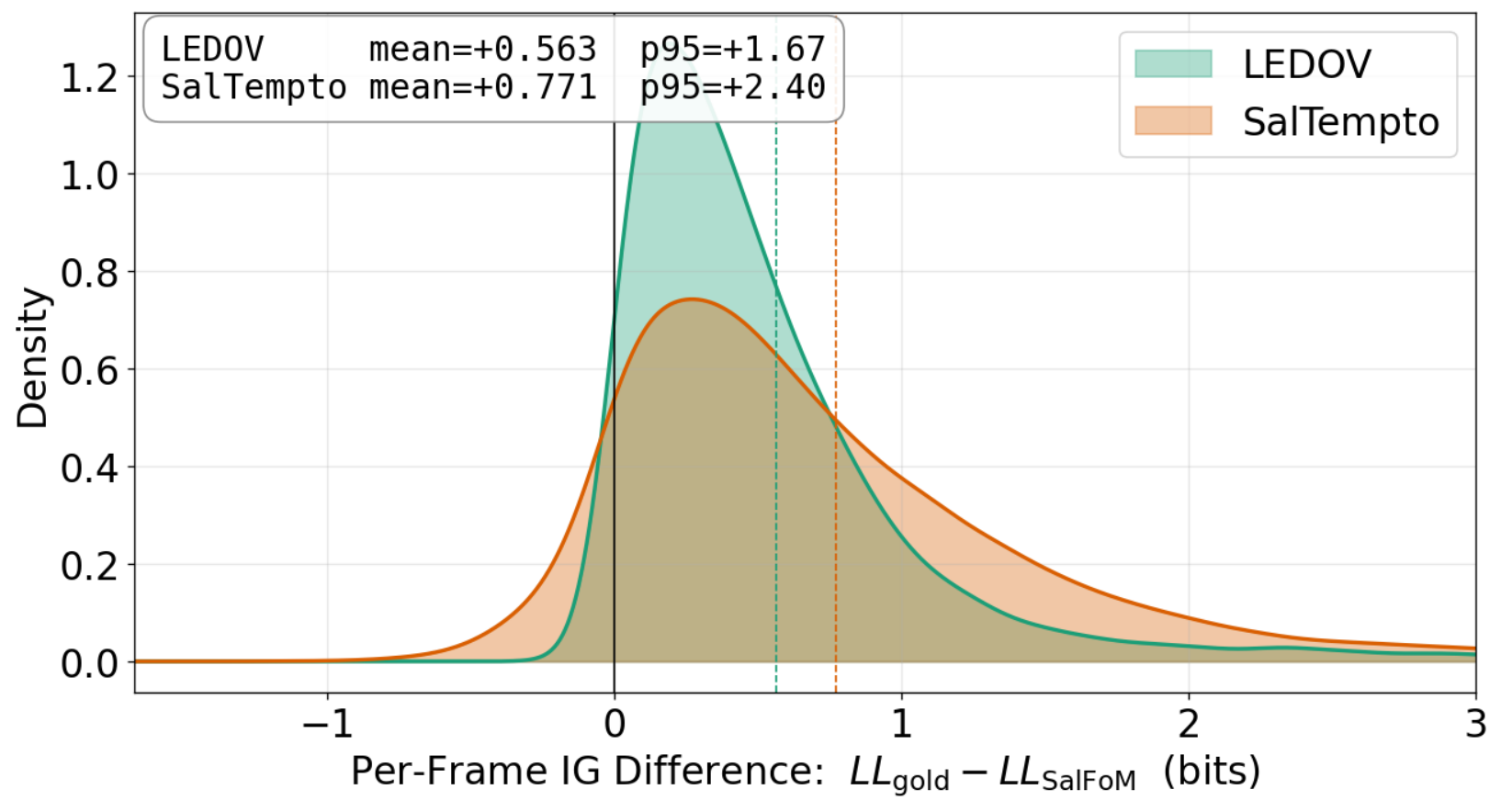}
  \caption{Per-frame residual distribution.}
  \label{fig:igdiff}
\end{subfigure}
\hfill
\begin{subfigure}[t]{0.36\linewidth}
  \centering
  \includegraphics[width=\linewidth]{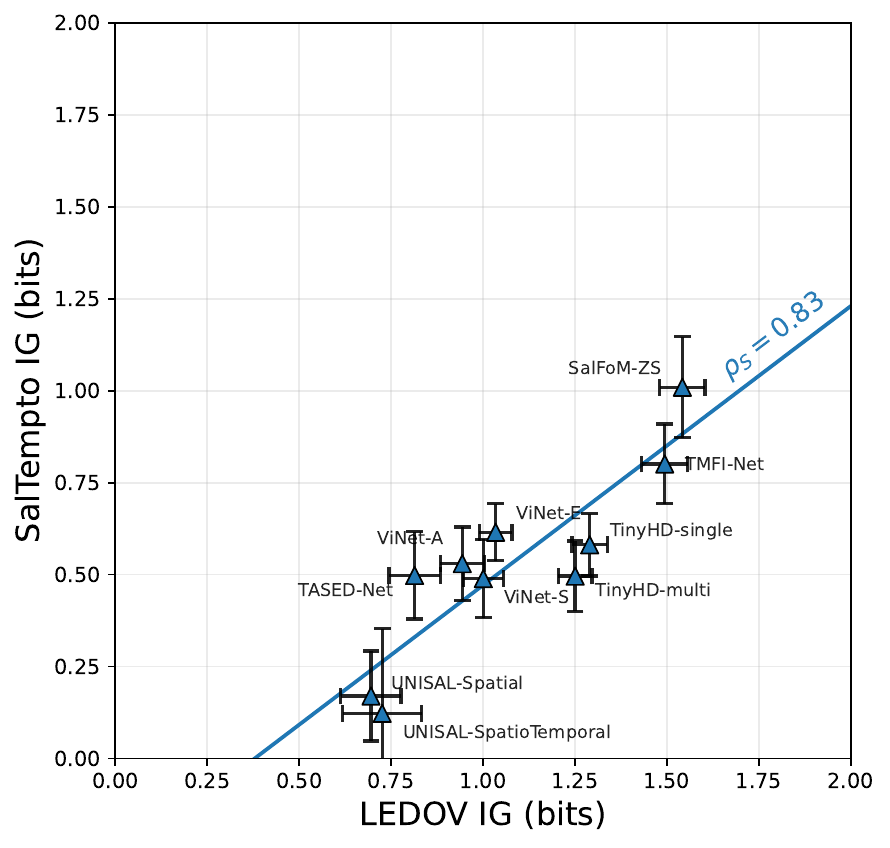}
  \caption{Cross-dataset IG.}
  \label{fig:cross_dataset_zeroshot}
\end{subfigure}
\caption{(a) Per-frame information-gain residual $\mathrm{LL}(gold) - \mathrm{LL}(SalFoM\text{-}Tuned)$ on LEDOV (green) vs \datasetname{} (orange). \datasetname{} is broader and less skewed, with a greater mean performance gap and a heavier tail. (b) Cross-dataset IG (bits) on 10 models evaluated in zero-shot on both LEDOV (x-axis) and \datasetname{} (y-axis); models trained or fine-tuned on either dataset are excluded to compute zero-shot performance rankings. Error bars are 95\,\% intervals for paired comparisons between models \citep{cousineau_confidence_2005,morey_confidence_2008}. Spearman $\rho = \lockedRhoExcluded{}$, implying model rankings are tightly preserved across datasets; the correlation including the LEDOV-pretrained models (zero-shot SalTempto) is $\rho = \lockedRho{}$ (App.~\ref{app:results-tables}).}
\label{fig:results}
\vspace{-2em}
\end{figure}

\xhdr{Cross-dataset rankings are largely consistent}
\label{sec:results:rankings}
We evaluate 10 video saliency models on LEDOV and SalTempto in zero-shot, finding Spearman $\rho = \lockedRhoExcluded{}$ between the LEDOV and \datasetname{} rankings pooled across val and test (Figure~\ref{fig:results}b; per-(dataset, split) tables in App.~\ref{app:results-tables}). The consistency mostly tracks architectural capability: SalFoM, a recent foundational video saliency model designed to exploit temporal patterns, holds rank 1 on each dataset; the UNISAL variants hold the bottom two ranks on each. The largest residual shifter is TinyHD-multi (rank 4 on LEDOV $\to$ rank 7 on \datasetname{}) with ViNet variants exchanging ranks within the model family. \datasetname{} and LEDOV exactly agree on the best and worst-performing model families.

\xhdr{\datasetname{} leaves more residual information to be explained than LEDOV}
\label{sec:results:residual}
Figure~\ref{fig:results}a plots the per-frame information-gain residual $\mathrm{LL}(p_{\mathrm{gold}}) - \mathrm{LL}(p_{\mathrm{SalFoM\text{-}Tuned}})$ on each dataset. The mean per-frame residual is larger on \datasetname{} (\lockedFrameMeanSalTempto{} vs \lockedFrameMeanLEDOV{}); the 95th-percentile residual is approximately 44\,\% larger (\lockedFrameTailSalTempto{} vs \lockedFrameTailLEDOV{}). The shape difference is informative: LEDOV has a high peak and a long thin right tail (skewness \lockedFrameSkewLEDOV{}), consistent with the ``mostly easy frames plus occasional outliers'' pattern that \citeauthor{tangemann_measuring_2020} described. \datasetname{} is broader and less skewed (\lockedFrameSkewSalTempto{}), with difficulty distributed across frames rather than concentrated in outliers.

\begin{figure}[t]
\centering
\includegraphics[width=0.9\linewidth]{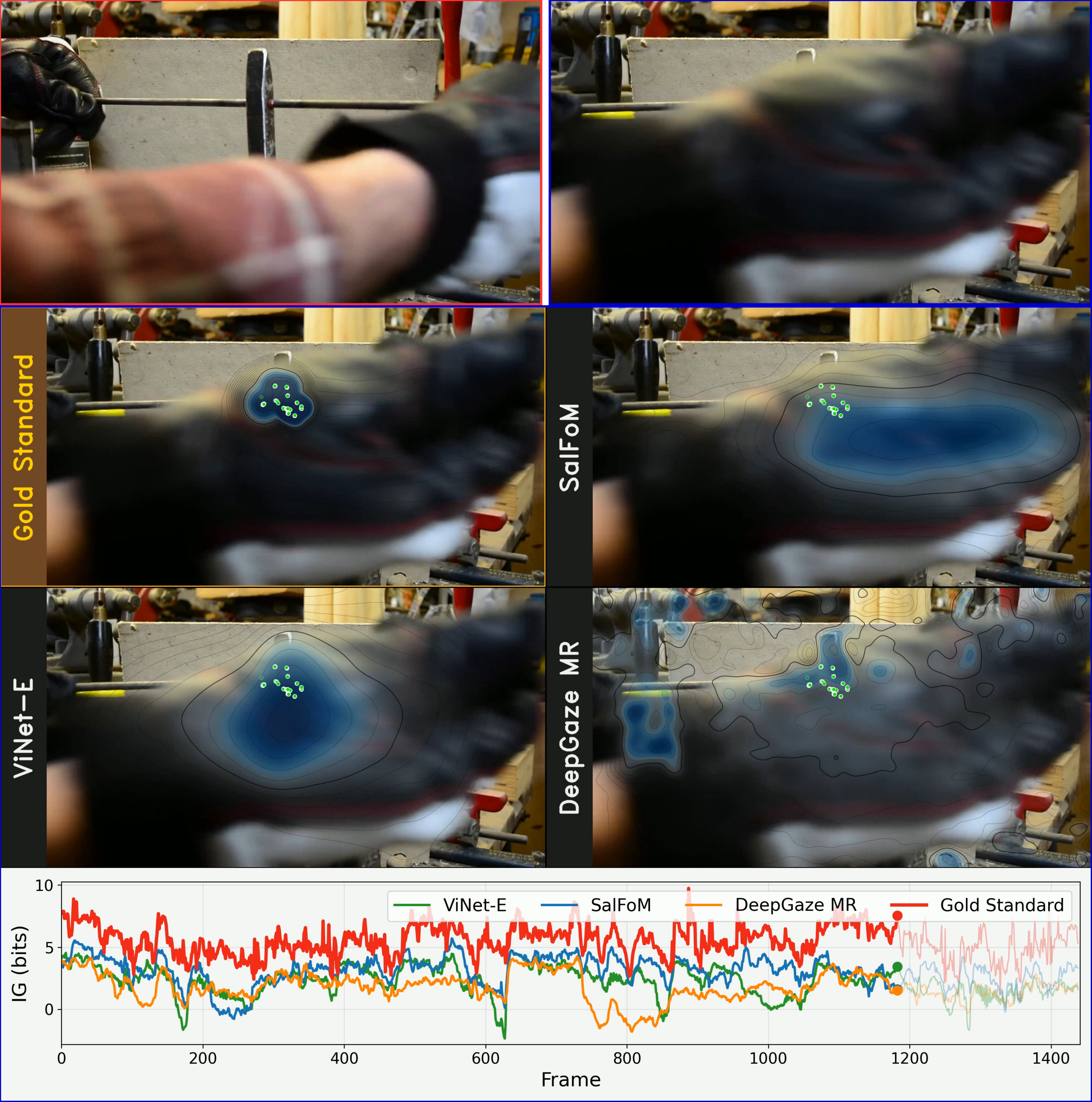}
\caption{\emph{Object permanence despite occlusion.} Workshop scene: A hot metal rod in a vice/stand (top-left, red border) gets transiently occluded by a gloved hand sweeping across (top-right, blue border). Gold-standard fixations remain locked at the rod's position despite occlusion. SalFoM redistributes mass onto the visible motion (the glove); ViNet-E concentrates centrally on the hand region; DeepGaze MR is diffuse across multiple salient elements. Generally, models attend to currently-visible salient content; subjects maintain attention on the (now-occluded) target.}
\label{fig:occlusion}
\end{figure}

\xhdr{Observed gaze behaviours}
\label{sec:results:failures}
What kinds of stimulus content drive the residual on \datasetname{}? \citet{tangemann_measuring_2020} reported, by visual inspection of static-baseline failure cases on LEDOV, that temporal effects on gaze placement exist but are rare and uncaptured by deep models. Their analysis identified three recurring patterns: \emph{interactions between objects}, where gaze concentrates at the interaction point rather than on the objects (e.g., a child feeding a giraffe); \emph{suddenly appearing objects}, which strongly attract gaze (e.g., text appearing on screen); and \emph{movements of objects}, where gaze tracks motion-bearing locations (e.g., a gymnast's torso during a swinging exercise).

We sample a few high-residual clips from the right tail of Figure~\ref{fig:results}a. Apart from the known failure modes discussed above (interactions, sudden object appearance, movement), which still show up, through manual examination of SalTempto videos and gaze recordings, we observe and characterise three additional visual behaviours that current models often fail to capture.

\emph{Object permanence despite occlusion \citep{churchland_evidence_2003,balaban_electrophysiology_2024}.} Subject gaze persists on a transiently occluded target, while models redistribute to currently visible salient content (Figure~\ref{fig:occlusion}). The corresponding video shows a workshop scene with an iron rod clamped on a vice. As observed by \citet{tangemann_measuring_2020}, the interaction point of the rod and vice accounts for a majority of the gaze density over a period of time. The figure shows that the exact region remains interesting for observers even when the point of interaction becomes occluded by a moving arm.

\begin{figure}[!ht]
\centering
\includegraphics[width=0.9\linewidth]{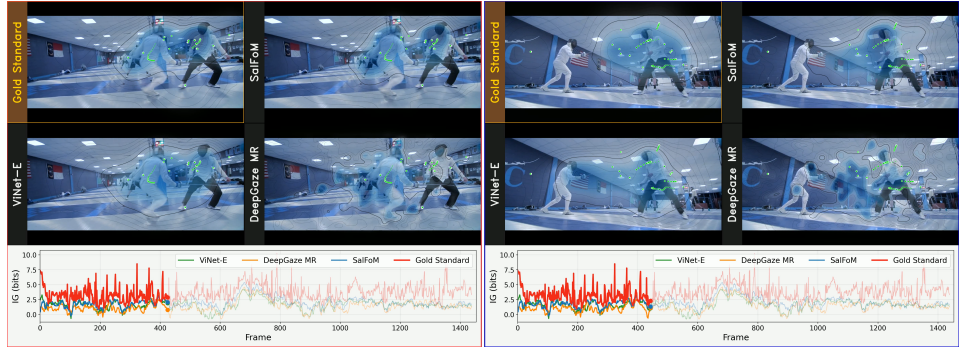}
\caption{\emph{Scene inertia.} The video cuts between different stances of two fencers. The first clip, pre-cut (left, red border), has both fencers at the right of the scene. Post-cut (right, blue border), the scene shows the fencers near the starting positions in a span of just a few frames. The human gaze readings (green dots) are highly consistent between the frames - the Gold Standard shows the highest densities at salient pre-cut regions that are non-salient post-cut. Deep Learning models quickly redistribute attention to the foreground figures based on current-frame salience. The effect is severe enough to perturb the gold standard itself due to its deep learning model component, so the residual-to-gold here underestimates the architectural shortfall by estimating high-variance low-peak densities.}
\label{fig:scene_inertia}
\vspace{-1em}
\end{figure}

\emph{Scene inertia \citep{carmi2006,wang_temporal_2012}.} Subject gaze persists across an abrupt scene-composition shift while models reattend to the new visual content. This can happen if there is a sudden change of the scene itself or if a sudden action occurs in an ongoing scene that causes a large displacement of a salient object in a short time, as shown in Figure~\ref{fig:scene_inertia}. The corresponding video cuts between two stances of a fencing bout: pre-cut, the fencers stand on the right of the frame; post-cut, within just a few frames, they have shifted to the starting positions. While human observers take a few seconds to reset their attention targets, models shift their predictions unrealistically fast. 

\begin{figure}[t]
\centering
\includegraphics[width=0.9\linewidth]{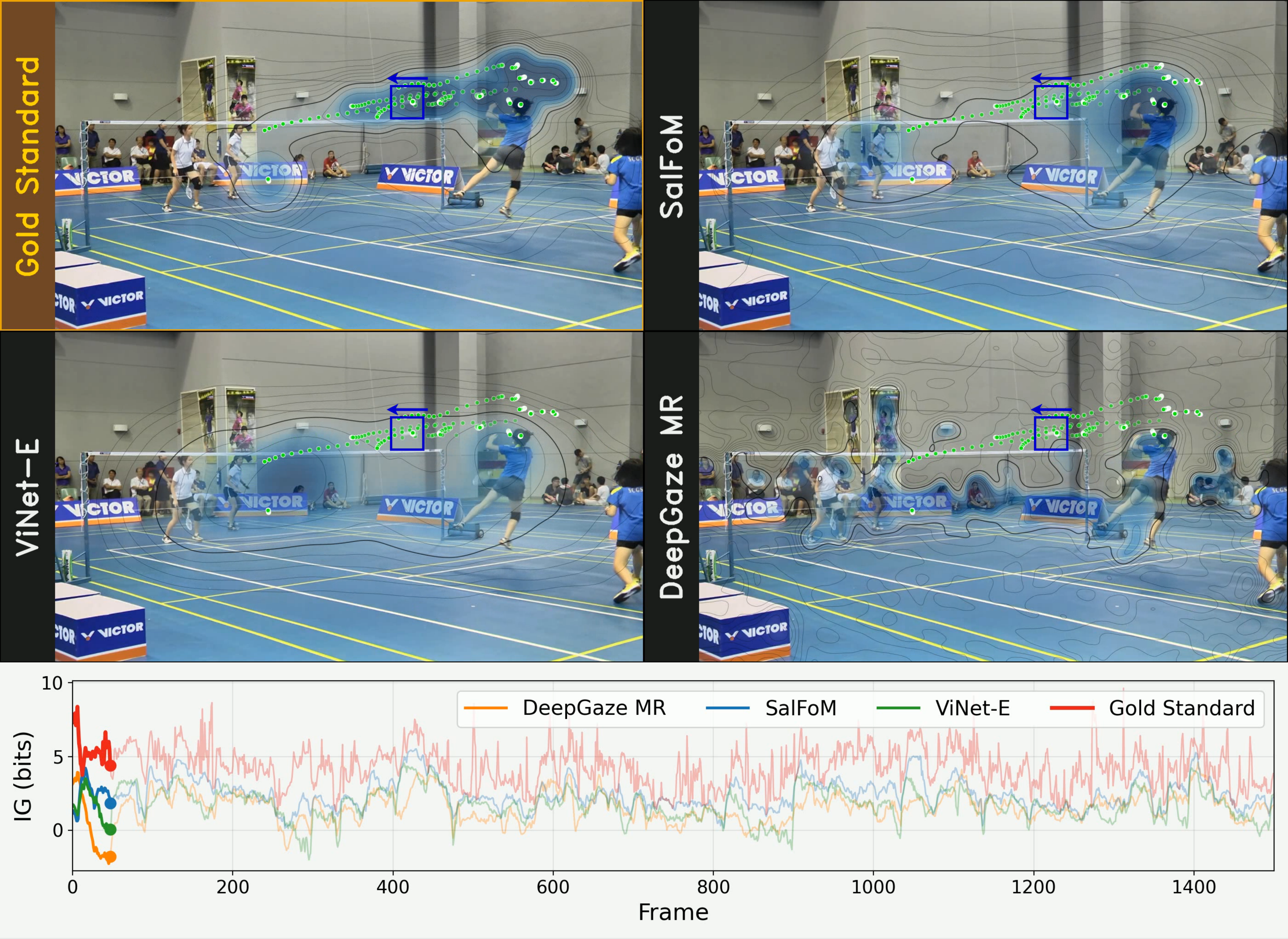}
\caption{\emph{Anticipatory saccades.} Badminton smash: shuttle moving right-to-left after the strike (blue arrows mark trajectory across frames, blue boxes locate shuttle). Subject fixations form a smooth trail spanning the shuttle's trajectory, with mass both trailing the shuttle (pursuit lag) \emph{and} leading it (anticipatory saccades). The models concentrate mass primarily at the players, missing the entity that viewers find most salient: the moving shuttle.}
\label{fig:anticipation}
\end{figure}

\emph{Anticipatory saccades \citep{vig_eye_2011,tonn_gaze_2025}.} Subject gaze leads moving content, pre-attending to where it will arrive, while models track it from current visual evidence and capture only the trailing component (Figure~\ref{fig:anticipation}). The corresponding video shows a badminton smash with the shuttle moving rapidly right-to-left across the frame after the strike. Subject fixations form a smooth trail spanning the shuttle's trajectory, with the leading edge of observer trajectories leading it (the anticipatory component). The models concentrate predicted gaze on the players and miss the moving shuttle that subjects find most salient.

\begin{figure}[t]
\centering
\includegraphics[trim={0 21cm 0 0},clip,width=0.9\linewidth]{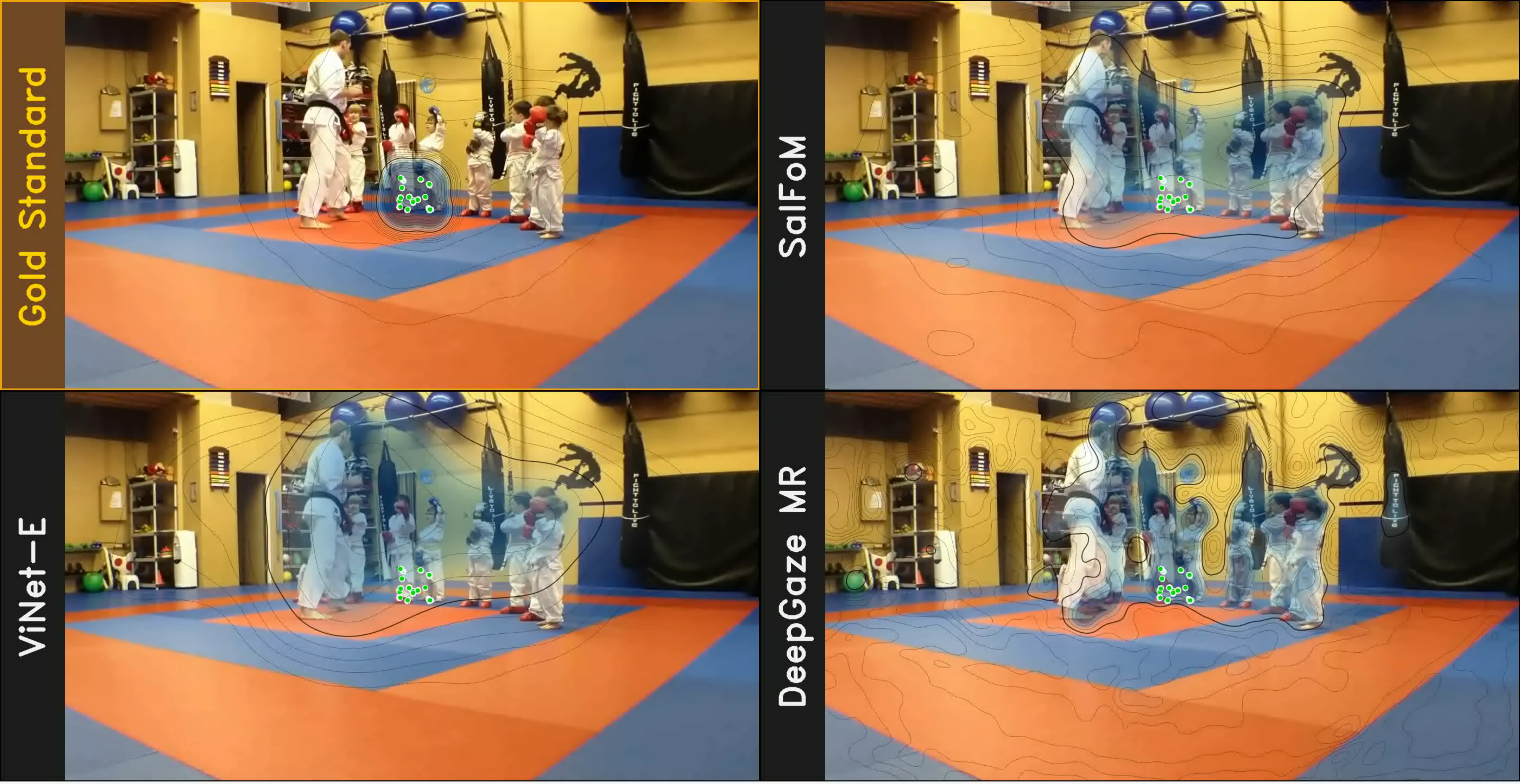}
\caption{\emph{Viewing-onset biases.} Karate dojo clip from \datasetname{} at the onset of the viewing session. Top-left: gold-standard density concentrates tightly at the center, reflecting the centre-anchored gaze that follows clip onset. Top-right and bottom row: SalFoM, ViNet-E, and DeepGaze MR distribute predicted attention across multiple visually salient agents and fail to capture scene-onset behaviour.}
\label{fig:early_frame}
\end{figure}



\section{Discussion}
\label{sec:discussion}


\xhdr{Architectural targets}
Each observed behaviour (\S\ref{sec:results:failures}) points at a structured prior that the current model class does not reliably encode: object permanence (occlusion), scene-state continuity (scene inertia), and event continuity (anticipatory saccades). Architectural implementations of each prior exist in adjacent fields, for example, memory-integrated architectures \citep{park2026videotitans,ravi2024sam2}, object state tracking \citep{ravi2024sam2,carion2025sam3segmentconcepts,Tokmakov2021LearningTT}, scene-cut detection \citep{soucek2020transnetv2,zhuautoshot}, motion estimation \citep{teed_raft_2020} and trajectory extrapolation \citep{trajpred} modules. These have not been applied to video saliency prediction widely, and open an interesting avenue for future research directions.

\xhdr{Self-inclusion in the gold standard}
\label{sec:discussion:framing}
The gold-standard mixture for each dataset includes SalFoM fine-tuned on that dataset as a mixture component. Substituting other panel models for this component changes neither the model ranking nor the cross-dataset contrast, and SalFoM recovers the smallest fraction of the headroom under its own ceiling (App.~\ref{app:gs-substitution}). In general a future model more capable than SalFoM can be substituted in for a better gold standard. The gold standard is, and should be, an improving target: as models improve, more of the available gaze information becomes measurable, raising the bar for the amount of information models are expected to explain.

\xhdr{Early-frame protocol}
\label{sec:discussion:earlyframe}
Subjects' gaze is centre-anchored for the first $\sim$15 frames following the fixation cross (Figure~\ref{fig:early_frame}). \citet{tangemann_measuring_2020} excluded these frames; we retain them because they are part of the empirical centerbias and excluding them would create a baseline-vs-model mismatch. We observe consistent model failure on the center-focused viewing pattern at video onset, but do not pool it with the previously discussed viewing behaviours, as this is specific to recording setups. We still show it as a failure case (Figure~\ref{fig:early_frame}): the data is readily available during training, so models should capture this behaviour.

\xhdr{Limitations}
\label{sec:discussion:limitations}
\datasetname{} is sourced from an action dataset, so its dynamics are those of human activity in the foreground, with camera motion and scene cuts as they occur in web video; content whose dynamics come from no acting agent is not targeted by the selection criterion, which limits the scope of the conclusions drawn. The evaluation splits comprise \Nval{} validation and \Ntest{} test videos; at one minute each they carry more viewing time than LEDOV's evaluation splits (\S\ref{sec:dataset:splits}) and span a broader range of content (\S\ref{sec:dataset:source}), and the static and temporal intervals do not overlap on either dataset (Table~\ref{tab:jackknife}). The dataset-level comparison is against LEDOV alone, for the reasons in \S\ref{sec:related}. 

\section{Conclusion}
\label{sec:conclusion}

\citet{tangemann_measuring_2020} diagnosed that LEDOV's predictable gaze structure is largely recoverable by a static baseline, and observed in passing that the temporal effects which do exist are rare and uncaptured by contemporary models. We have shown that the diagnosis remains valid under a panel of recent state-of-the-art architectures: a fine-tuned static probe still recovers more than half of LEDOV's headroom above centerbias, and the strongest current temporal model improves on it only modestly. \datasetname{} is the stand-alone benchmark with greater dynamism this diagnosis implies. The same fine-tuned static probe recovers only a small fraction of \datasetname{}'s headroom; even with dataset-specific fine-tuning, the strongest temporal model in the panel leaves nearly half of it unexplained, and shows a heavier-tailed per-frame residual than on LEDOV. LEDOV is still useful for ranking general saliency architectures, but it is weak as a diagnostic for temporal attention - the key aspect that is supposed to distinguish the video saliency problem from static image saliency modelling. SalTempto is a more temporally demanding benchmark that reveals both gains from current temporal models and substantial remaining failures. The three gaze behaviours we describe (\emph{object permanence despite occlusion}, \emph{scene inertia}, and \emph{anticipatory saccades}), alongside the failure modes categorised by \citet{tangemann_measuring_2020}, characterise where the remaining gap lies; how often each occurs and how much of the residual each explains requires a dedicated annotation study, which we leave to future work. Our results indicate that the field has measured architectural progress against a benchmark that does not measurably require temporal reasoning, and we provide a setting on which the field can begin to address this mismatch directly.


\section*{AI use statement}
In this work, we used generative AI tools for some of the tasks with required disclosure: coding agents were used in part to assist with implementation, mainly in setting up coding environments and automating replication of public baselines from GitHub; the remaining required-disclosure tasks (synthetic data generation, theoretical or mathematical claims and proofs, translation, qualitative or thematic data analysis) are not applicable to this work. Additionally, we used generative AI tools for tasks with recommended disclosure: to edit the research paper to improve readability, to draft parts of the paper from author-written notes, to create or edit software code (\LaTeX{} and figure- and table-formatting scripts). We have reviewed all AI-assisted work: all AI-assisted text was read and revised by the authors, all AI-assisted code was checked by the authors. All numbers were taken from evaluation script outputs and not fabricated or hallucinated by AI agents. We take responsibility for the final content of this work, including text, claims or artifacts produced with the aid of generative AI.
AI agents were not used to propose or refine hypotheses, to design or provide feedback on the research methodology or experiments, to clean or reformat the dataset, or to interpret results.

\section*{Ethics statement}
\datasetname{} involves research with human subjects. The eye-tracking experiment was approved by the institutional ethics board (institution withheld for anonymity), and informed consent was obtained from all participants prior to recording (\S\ref{sec:dataset:eye}). The procedures (chin-rest stabilisation, 5-point calibration, drift checks between blocks, self-paced breaks) follow standard psychophysics practice with no known risks beyond brief screen viewing. Released gaze data are anonymised. The stimuli are derived from the publicly available HACS-Segments dataset \citep{zhao_hacs_2019}; \datasetname{} will be released as a hosted benchmark on acceptance (\S\ref{sec:dataset:splits}). Video saliency prediction is a non-generative, non-decision-making research area, and we do not see a direct pathway to misuse.

\section*{Reproducibility statement}
Section~\ref{sec:methods} documents the stimulus selection criteria, the recording apparatus and protocol, the train/validation/test splits, the model panel with the provenance of each released checkpoint, the fine-tuning protocol and its compute footprint, and the construction of the centerbias and gold-standard mixture. App.~\ref{app:results-tables} defines the IGE metric and reports LL, IGE, KLD, CC, SIM, AUC and NSS for every model on every (dataset, split) pair, so that any number in the main text can be traced to a table cell. App.~\ref{app:dynamism} specifies how the dataset-level dynamism metrics are computed. An anonymised repository (\url{https://huggingface.co/datasets/bethgelab/video_saliency}) holds the training and validation videos with the underlying raw 500\,Hz gaze trajectories and per-frame binned positions, the test videos with gaze held out, and the evaluation code; the hosted benchmark is released on acceptance (\S\ref{sec:dataset:splits}). All benchmarked architectures are evaluated from their publicly released checkpoints, cited in \S\ref{sec:methods:framework}.

\section*{Acknowledgements}
Susmit Agrawal was supported by the German Research Foundation (DFG):
SFB 1233, Robust Vision: Inference Principles and Neural Mechanisms, TP C2,
project number: 276693517. Julianne Verwiebe is supported by the Carl Zeiss Foundation through the project “Certification and Foundations of Safe ML Systems”. 
Both authors thank the International Max Planck Research School for Intelligent Systems (IMPRS-IS) for support. All authors were supported by the Tübingen AI Center.


\clearpage
\newpage

\bibliography{TemporalSaliencyBenchmarking}

@inproceedings{tangemann_measuring_2020,
	address = {Cham},
	title = {Measuring the {Importance} of {Temporal} {Features} in {Video} {Saliency}},
	isbn = {978-3-030-58604-1},
	doi = {10.1007/978-3-030-58604-1_40},
	language = {en},
	booktitle = {Computer {Vision} – {ECCV} 2020},
	publisher = {Springer International Publishing},
	author = {Tangemann, Matthias and Kümmerer, Matthias and Wallis, Thomas S. A. and Bethge, Matthias},
	editor = {Vedaldi, Andrea and Bischof, Horst and Brox, Thomas and Frahm, Jan-Michael},
	year = {2020},
	pages = {667--684},
}

@misc{zhao_hacs_2019,
	title = {{HACS}: {Human} {Action} {Clips} and {Segments} {Dataset} for {Recognition} and {Temporal} {Localization}},
	shorttitle = {{HACS}},
	url = {http://arxiv.org/abs/1712.09374},
	doi = {10.48550/arXiv.1712.09374},
	urldate = {2025-04-09},
	publisher = {arXiv},
	author = {Zhao, Hang and Torralba, Antonio and Torresani, Lorenzo and Yan, Zhicheng},
	month = sep,
	year = {2019},
	note = {arXiv:1712.09374 [cs]},
}

@article{mital_clustering_2011,
	title = {Clustering of {Gaze} {During} {Dynamic} {Scene} {Viewing} is {Predicted} by {Motion}},
	volume = {3},
	copyright = {http://www.springer.com/tdm},
	issn = {1866-9956, 1866-9964},
	url = {http://link.springer.com/10.1007/s12559-010-9074-z},
	doi = {10.1007/s12559-010-9074-z},
	language = {en},
	number = {1},
	urldate = {2025-04-12},
	journal = {Cognitive Computation},
	author = {Mital, Parag K. and Smith, Tim J. and Hill, Robin L. and Henderson, John M.},
	month = mar,
	year = {2011},
	pages = {5--24},
}

@misc{wang_revisiting_2018,
	title = {Revisiting {Video} {Saliency}: {A} {Large}-scale {Benchmark} and a {New} {Model}},
	shorttitle = {Revisiting {Video} {Saliency}},
	url = {http://arxiv.org/abs/1801.07424},
	doi = {10.48550/arXiv.1801.07424},
	language = {en},
	urldate = {2025-04-15},
	publisher = {arXiv},
	author = {Wang, Wenguan and Shen, Jianbing and Guo, Fang and Cheng, Ming-Ming and Borji, Ali},
	month = may,
	year = {2018},
	note = {arXiv:1801.07424 [cs]},
}

@article{mathe_actions_2015,
	title = {Actions in the {Eye}: {Dynamic} {Gaze} {Datasets} and {Learnt} {Saliency} {Models} for {Visual} {Recognition}},
	volume = {37},
	issn = {1939-3539},
	shorttitle = {Actions in the {Eye}},
	url = {https://ieeexplore.ieee.org/document/6942210},
	doi = {10.1109/TPAMI.2014.2366154},
	number = {7},
	urldate = {2025-04-15},
	journal = {IEEE Transactions on Pattern Analysis and Machine Intelligence},
	author = {Mathe, Stefan and Sminchisescu, Cristian},
	month = jul,
	year = {2015},
	pages = {1408--1424},
}

@incollection{jiang_predicting_2018,
	title = {Predicting {Video} {Saliency} with {Object}-to-{Motion} {CNN} and {Two}-layer {Convolutional} {LSTM}},
	booktitle = {Computer {Vision} -- {ECCV} 2018},
	publisher = {Springer},
	volume = {11218},
	url = {http://arxiv.org/abs/1709.06316},
	urldate = {2025-04-15},
	author = {Jiang, Lai and Xu, Mai and Wang, Zulin},
	year = {2018},
	doi = {10.1007/978-3-030-01264-9_37},
	note = {arXiv:1709.06316 [cs]},
	pages = {625--642},
}

@article{bylinskii_what_2019,
	title = {What do different evaluation metrics tell us about saliency models?},
	volume = {41},
	number = {3},
	pages = {740--757},
	year = {2019},
	journal = {IEEE Transactions on Pattern Analysis and Machine Intelligence},
	author = {Bylinskii, Zoya and Judd, Tilke and Oliva, Aude and Torralba, Antonio and Durand, Fr\'{e}do},
	publisher = {IEEE},
	doi = {10.1109/TPAMI.2018.2815601},
}

@article{kummerer_information-theoretic_2015,
	title = {Information-theoretic model comparison unifies saliency metrics},
	volume = {112},
	url = {https://www.pnas.org/doi/abs/10.1073/pnas.1510393112},
	doi = {10.1073/pnas.1510393112},
	language = {EN},
	number = {52},
	urldate = {2025-04-16},
	journal = {Proceedings of the National Academy of Sciences},
	author = {Kümmerer, Matthias and Wallis, Thomas S. A. and Bethge, Matthias},
	month = dec,
	year = {2015},
	note = {Company: National Academy of Sciences
Distributor: National Academy of Sciences
Institution: National Academy of Sciences
Label: National Academy of Sciences
Publisher: Proceedings of the National Academy of Sciences},
	pages = {16054--16059},
}

@misc{kummerer_deepgaze_2016,
	title = {{DeepGaze} {II}: {Reading} fixations from deep features trained on object recognition},
	shorttitle = {{DeepGaze} {II}},
	url = {http://arxiv.org/abs/1610.01563},
	doi = {10.48550/arXiv.1610.01563},
	urldate = {2025-04-27},
	publisher = {arXiv},
	author = {Kümmerer, Matthias and Wallis, Thomas S. A. and Bethge, Matthias},
	month = oct,
	year = {2016},
	note = {arXiv:1610.01563 [cs]},
}

@inproceedings{min_tasednet_2019,
  title = {{TASED}-{Net}: {Temporally}-{Aggregating} {Spatial} {Encoder}-{Decoder} {Network} for {Video} {Saliency} {Detection}},
  booktitle = {Proceedings of the {IEEE}/{CVF} {International} {Conference} on {Computer} {Vision} ({ICCV})},
  author = {Min, Kyle and Corso, Jason J.},
  year = {2019},
  pages = {2394--2403},
}

@inproceedings{jain_vinet_2021,
  title = {{ViNet}: {Pushing} the {Limits} of {Visual} {Modality} for {Audio}-{Visual} {Saliency} {Prediction}},
  booktitle = {2021 {IEEE}/{RSJ} {International} {Conference} on {Intelligent} {Robots} and {Systems} ({IROS})},
  author = {Jain, Samyak and Yarlagadda, Pradeep and Jyoti, Shreyank and Karthik, Shyamgopal and Subramanian, Ramanathan and Gandhi, Vineet},
  year = {2021},
  pages = {3520--3527},
  doi = {10.1109/IROS51168.2021.9635989},
}

@article{zhou_tmfinet_2023,
  title = {Transformer-{Based} {Multi}-{Scale} {Feature} {Integration} {Network} for {Video} {Saliency} {Prediction}},
  journal = {IEEE Transactions on Circuits and Systems for Video Technology},
  author = {Zhou, Xiaofei and Wu, Songhe and Shi, Ran and Zheng, Bolun and Wang, Shuai and Yin, Haibing and Zhang, Jiyong and Yan, Chenggang},
  year = {2023},
  volume = {33},
  number = {12},
  pages = {7696--7707},
  doi = {10.1109/TCSVT.2023.3278410},
}

@inproceedings{moradi_salfom_2024,
  title = {{SalFoM}: {Dynamic} {Saliency} {Prediction} with {Video} {Foundation} {Models}},
  booktitle = {Proceedings of the {International} {Conference} on {Pattern} {Recognition} ({ICPR})},
  author = {Moradi, Morteza and Rundo, Francesco and Spampinato, Concetto and Borji, Ali and Palazzo, Simone},
  year = {2024},
  eprint = {2404.03097},
  archivePrefix = {arXiv},
  primaryClass = {cs.CV},
}

@inproceedings{teed_raft_2020,
  title = {{RAFT}: {Recurrent} {All}-{Pairs} {Field} {Transforms} for {Optical} {Flow}},
  booktitle = {European {Conference} on {Computer} {Vision} ({ECCV})},
  author = {Teed, Zachary and Deng, Jia},
  year = {2020},
  pages = {402--419},
  doi = {10.1007/978-3-030-58536-5_24},
}

@article{kummerer_predicting_2023,
  title = {Predicting {Visual} {Fixations}},
  author = {K{\"u}mmerer, Matthias and Bethge, Matthias},
  year = {2023},
  journal = {Annual Review of Vision Science},
  volume = {9},
  pages = {269--291},
  doi = {10.1146/annurev-vision-120822-072528},
}

@inproceedings{hu_tinyhd_2023,
  title = {{TinyHD}: {Efficient} {Video} {Saliency} {Prediction} with {Heterogeneous} {Decoders} using {Hierarchical} {Maps} {Distillation}},
  booktitle = {Proceedings of the {IEEE}/{CVF} {Winter} {Conference} on {Applications} of {Computer} {Vision} ({WACV})},
  author = {Hu, Feiyan and Palazzo, Simone and {Proietto Salanitri}, Federica and Bellitto, Giovanni and Moradi, Morteza and Spampinato, Concetto and McGuinness, Kevin},
  year = {2023},
  pages = {2051--2060},
}

@inproceedings{droste_unisal_2020,
  title = {Unified {Image} and {Video} {Saliency} {Modeling}},
  booktitle = {European {Conference} on {Computer} {Vision} ({ECCV})},
  author = {Droste, Richard and Jiao, Jianbo and Noble, J. Alison},
  year = {2020},
  pages = {419--435},
  doi = {10.1007/978-3-030-58558-7_25},
}

@inproceedings{li_uniformer_2022,
  title = {{UniFormer}: {Unified} {Transformer} for {Efficient} {Spatiotemporal} {Representation} {Learning}},
  booktitle = {International {Conference} on {Learning} {Representations} ({ICLR})},
  author = {Li, Kunchang and Wang, Yali and Gao, Peng and Song, Guanglu and Liu, Yu and Li, Hongsheng and Qiao, Yu},
  year = {2022},
  eprint = {2201.04676},
  archivePrefix = {arXiv},
}

@misc{agrawal2026raisingceilingbetterempirical,
      title={Raising the Ceiling: Better Empirical Fixation Densities for Saliency Benchmarking}, 
      author={Susmit Agrawal and Jannis Hollman and Matthias Kümmerer},
      year={2026},
      eprint={2605.03885},
      archivePrefix={arXiv},
      primaryClass={cs.CV},
      url={https://arxiv.org/abs/2605.03885}, 
}

@inproceedings{
park2026videotitans,
title={VideoTitans: Scalable Video Prediction with Integrated Short- and Long-term Memory},
author={Young-Jae Park and Minseok Seo and Hae-Gon Jeon},
booktitle={The Thirty-ninth Annual Conference on Neural Information Processing Systems},
year={2025},
url={https://openreview.net/forum?id=86enCXORIV}
}

@article{ravi2024sam2,
  title={SAM 2: Segment Anything in Images and Videos},
  author={Ravi, Nikhila and Gabeur, Valentin and Hu, Yuan-Ting and Hu, Ronghang and Ryali, Chaitanya and Ma, Tengyu and Khedr, Haitham and R{\"a}dle, Roman and Rolland, Chloe and Gustafson, Laura and Mintun, Eric and Pan, Junting and Alwala, Kalyan Vasudev and Carion, Nicolas and Wu, Chao-Yuan and Girshick, Ross and Doll{\'a}r, Piotr and Feichtenhofer, Christoph},
  journal={arXiv preprint arXiv:2408.00714},
  url={https://arxiv.org/abs/2408.00714},
  year={2024}
}

@misc{carion2025sam3segmentconcepts,
      title={SAM 3: Segment Anything with Concepts},
      author={Nicolas Carion and Laura Gustafson and Yuan-Ting Hu and Shoubhik Debnath and Ronghang Hu and Didac Suris and Chaitanya Ryali and Kalyan Vasudev Alwala and Haitham Khedr and Andrew Huang and Jie Lei and Tengyu Ma and Baishan Guo and Arpit Kalla and Markus Marks and Joseph Greer and Meng Wang and Peize Sun and Roman Rädle and Triantafyllos Afouras and Effrosyni Mavroudi and Katherine Xu and Tsung-Han Wu and Yu Zhou and Liliane Momeni and Rishi Hazra and Shuangrui Ding and Sagar Vaze and Francois Porcher and Feng Li and Siyuan Li and Aishwarya Kamath and Ho Kei Cheng and Piotr Dollár and Nikhila Ravi and Kate Saenko and Pengchuan Zhang and Christoph Feichtenhofer},
      year={2025},
      eprint={2511.16719},
      archivePrefix={arXiv},
      primaryClass={cs.CV},
      url={https://arxiv.org/abs/2511.16719},
}

@article{Tokmakov2021LearningTT,
  title={Learning to Track with Object Permanence},
  author={Pavel Tokmakov and Jie Li and Wolfram Burgard and Adrien Gaidon},
  journal={2021 IEEE/CVF International Conference on Computer Vision (ICCV)},
  year={2021},
  pages={10840-10849},
  url={https://api.semanticscholar.org/CorpusID:232380249}
}

@article{vig_eye_2011,
	title = {Eye {Movements} {Show} {Optimal} {Average} {Anticipation} with {Natural} {Dynamic} {Scenes}},
	volume = {3},
	issn = {1866-9964},
	url = {https://doi.org/10.1007/s12559-010-9061-4},
	doi = {10.1007/s12559-010-9061-4},
	number = {1},
	journal = {Cognitive Computation},
	author = {Vig, Eleonora and Dorr, Michael and Martinetz, Thomas and Barth, Erhardt},
	month = mar,
	year = {2011},
	pages = {79--88},
}

@article{tonn_gaze_2025,
	title = {A gaze into the void: {Anticipatory} saccades toward prevented events},
	volume = {87},
	issn = {1943-393X},
	url = {https://doi.org/10.3758/s13414-025-03019-4},
	doi = {10.3758/s13414-025-03019-4},
	number = {3},
	journal = {Attention, Perception, \& Psychophysics},
	author = {Tonn, Solveig and Teigeler, Janna and Pfister, Roland and Gamer, Matthias},
	month = apr,
	year = {2025},
	pages = {848--861},
}

@inproceedings{carmi2006,
author = {Carmi, Ran and Itti, Laurent},
title = {Causal saliency effects during natural vision},
year = {2006},
isbn = {1595933050},
publisher = {Association for Computing Machinery},
address = {New York, NY, USA},
url = {https://doi.org/10.1145/1117309.1117313},
doi = {10.1145/1117309.1117313},
booktitle = {Proceedings of the 2006 Symposium on Eye Tracking Research \& Applications},
pages = {11–18},
numpages = {8},
location = {San Diego, California},
series = {ETRA '06}
}

@article{wang_temporal_2012,
	address = {United States},
	title = {Temporal eye movement strategies during naturalistic viewing.},
	volume = {12},
	issn = {1534-7362},
	doi = {10.1167/12.1.16},
	language = {eng},
	number = {1},
	journal = {Journal of vision},
	author = {Wang, Helena X. and Freeman, Jeremy and Merriam, Elisha P. and Hasson, Uri and Heeger, David J.},
	month = jan,
	year = {2012},
	pages = {16},
}

@article{churchland_evidence_2003,
	address = {United States},
	title = {Evidence for object permanence in the smooth-pursuit eye movements of monkeys.},
	volume = {90},
	issn = {0022-3077 1522-1598},
	doi = {10.1152/jn.01056.2002},
	language = {eng},
	number = {4},
	journal = {Journal of neurophysiology},
	author = {Churchland, Mark M. and Chou, I.-Han and Lisberger, Stephen G.},
	month = oct,
	year = {2003},
	pages = {2205--2218},
}

@article{balaban_electrophysiology_2024,
	title = {Electrophysiology {Reveals} {That} {Intuitive} {Physics} {Guides} {Visual} {Tracking} and {Working} {Memory}},
	volume = {8},
	issn = {2470-2986},
	url = {https://doi.org/10.1162/opmi_a_00174},
	doi = {10.1162/opmi_a_00174},
	journal = {Open Mind},
	author = {Balaban, Halely and Smith, Kevin A. and Tenenbaum, Joshua B. and Ullman, Tomer D.},
	month = nov,
	year = {2024},
	note = {\_eprint: https://direct.mit.edu/opmi/article-pdf/doi/10.1162/opmi\_a\_00174/2483508/opmi\_a\_00174.pdf},
	pages = {1425--1446},
}

@article{soucek2020transnetv2,
    title={TransNet V2: An effective deep network architecture for fast shot transition detection},
    author={Sou{\v{c}}ek, Tom{\'a}{\v{s}} and Loko{\v{c}}, Jakub},
    year={2020},
    journal={arXiv preprint arXiv:2008.04838},
}

@inproceedings{zhuautoshot,
  title={AutoShot: A Short Video Dataset and State-of-the-Art Shot Boundary Detection},
  author={Zhu, Wentao and Huang, Yufang and Xie, Xiufeng and Liu, Wenxian and Deng, Jincan and Zhang, Debing and Wang, Zhangyang and Liu, Ji},
  booktitle={Proceedings of the IEEE/CVF Conference on Computer Vision and Pattern Recognition Workshops (CVPRW)},
  year={2023}
}

@inproceedings{trajpred,
  author={Zhao, Yang and Du, Jingyu and Deng, Ruoyu and Cheng, Hong},
  booktitle={2024 IEEE Intelligent Vehicles Symposium (IV)}, 
  title={Target-point Attention Transformer: A novel trajectory predict network for end-to-end autonomous driving}, 
  year={2024},
  volume={},
  number={},
  pages={2325-2330},
  doi={10.1109/IV55156.2024.10588617}}

@article{cousineau_confidence_2005,
	title = {Confidence intervals in within-subject designs: {A} simpler solution to {Loftus} and {Masson}'s method},
	volume = {1},
	number = {1},
	journal = {Tutorials in Quantitative Methods for Psychology},
	author = {Cousineau, Denis},
	year = {2005},
	pages = {42--45},
	doi = {10.20982/tqmp.01.1.p042},
}

@article{morey_confidence_2008,
	title = {Confidence intervals from normalized data: {A} correction to {Cousineau} (2005)},
	volume = {4},
	number = {2},
	journal = {Tutorials in Quantitative Methods for Psychology},
	author = {Morey, Richard D.},
	year = {2008},
	pages = {61--64},
	doi = {10.20982/tqmp.04.2.p061},
}

@InProceedings{agrawal2026deepgaze3.5,
author="Agrawal, Susmit
and Bethge, Matthias
and K{\"u}mmerer, Matthias",
editor="Favaro, Paolo
and Kukelova, Zuzana
and Maki, Atsuto
and Rohrbach, Anna
and Schindler, Konrad
and Tombari, Federico",
title="DeepGaze3.5-VL: Modeling Scanpaths via Autoregressive Token Prediction",
booktitle="Computer Vision -- ECCV 2026",
year="2026",
publisher="Springer Nature Switzerland",
address="Cham",
pages="242--269",
isbn="978-3-032-37271-0"
}

@ARTICLE{schwetlick2024potsdam,
    
AUTHOR={Schwetlick, Lisa  and Kümmerer, Matthias  and Bethge, Matthias  and Engbert, Ralf },
           
TITLE={Potsdam data set of eye movement on natural scenes (DAEMONS)},
          
JOURNAL={Frontiers in Psychology},
          
VOLUME={Volume 15 - 2024},
  
YEAR={2024},
  
URL={https://www.frontiersin.org/journals/psychology/articles/10.3389/fpsyg.2024.1389609},
  
DOI={10.3389/fpsyg.2024.1389609},
  
ISSN={1664-1078},
  
}
\bibliographystyle{iclr2027_conference}


\clearpage
\newpage

\appendix

\section{Per-(dataset, split) Results}
\label{app:results-tables}

We follow the information-theoretic evaluation framework of \citet{kummerer_information-theoretic_2015,kummerer_predicting_2023}: models are scored by the log-likelihood (LL) they assign to held-out gaze. To ground LL against a known baseline, the Information Gain (IG) with respect to the dataset's centerbias is computed as $\mathrm{IG} = \mathrm{LL}_{\mathrm{model}} - \mathrm{LL}_{\mathrm{CB}}$, with the gold standard's IG defining an empirical upper bound $\mathrm{IG}_{\mathrm{gold}} = \mathrm{LL}_{\mathrm{gold}} - \mathrm{LL}_{\mathrm{CB}}$. This allows us to quantify how much of the gap between the centerbias and the gold standard is captured by a model, through the ratio of the two quantities, defined as the Information Gain Explained (IGE). By construction $\mathrm{IGE}(p_{\mathrm{CB}}) = 0$ and $\mathrm{IGE}(p_{\mathrm{gold}}) = 1$:
\begin{equation}
\mathrm{IGE}(\hat{p}) = \frac{\mathrm{LL}(\hat{p}) - \mathrm{LL}(p_{\mathrm{CB}})}{\mathrm{LL}(p_{\mathrm{gold}}) - \mathrm{LL}(p_{\mathrm{CB}})}.
\label{eq:ige}
\end{equation}

\begin{table}[ht]
\centering
\small
\caption{IGE (Eq.~\ref{eq:ige}) for the static baseline (DG-MR) and the strongest temporal model (SalFoM) on each dataset's validation and test splits post fine-tuning. SalFoM is fine-tuned by us on each dataset's training split. DG-MR is fine-tuned by us on \datasetname{} and uses the released LEDOV-trained checkpoint from \citet{tangemann_measuring_2020} on LEDOV.}
\label{tab:matched_protocol}
\begin{tabular}{lcccc}
\toprule
Model & \multicolumn{2}{c}{LEDOV} & \multicolumn{2}{c}{\datasetname{}} \\
\cmidrule(lr){2-3} \cmidrule(lr){4-5}
      & val & test & val & test \\
\midrule
DG-MR  & \lockedLEDOVrecovVal{} & \lockedLEDOVrecovTest{} & \lockedSalTemptoRecovVal{} & \lockedSalTemptoRecovTest{} \\
SalFoM & \lockedLEDOVbestIGE{} & \lockedLEDOVbestIGEtest{} & \lockedSalTemptoBestIGEval{} & \lockedSalTemptoBestIGEtest{} \\
\bottomrule
\end{tabular}
\end{table}

Table~\ref{tab:jackknife} reports the same quantities pooled over all evaluation videos of each dataset, with leave-one-video-out jackknife intervals.
\begin{table}[ht]
\centering
\small
\caption{Fine-tuned results with leave-one-video-out jackknife intervals. IG is the information gain over the centerbias in bits per fixation and IGE the fraction of the centerbias-to-gold headroom recovered, both pooled over the validation and test videos of each dataset (77 videos on LEDOV, 20 on \datasetname{}) with every video weighted equally. Intervals are 95\,\% jackknife intervals over videos. Gap rows give SalFoM minus DG-MR.}
\label{tab:jackknife}
\begin{tabular}{llrlrl}
\toprule
Dataset & Model & \multicolumn{2}{c}{IG (bits)} & \multicolumn{2}{c}{IGE (\%)} \\
\cmidrule(lr){3-4} \cmidrule(lr){5-6}
 & & mean & 95\,\% jackknife & mean & 95\,\% jackknife \\
\midrule
LEDOV & Gold standard & 2.359 & [2.210, 2.507] & 100.0 & --- \\
 & DG-MR (static) & 1.321 & [1.170, 1.472] & 56.0 & [51.6, 60.4] \\
 & SalFoM (temporal) & 1.775 & [1.631, 1.918] & 75.2 & [72.6, 77.9] \\
 & Gap (SalFoM $-$ DG-MR) & 0.453 & [0.371, 0.535] & 19.2 & [15.8, 22.7] \\
\midrule
\datasetname{} & Gold standard & 1.864 & [1.613, 2.115] & 100.0 & --- \\
 & DG-MR (static) & 0.243 & [0.011, 0.474] & 13.0 & [0.8, 25.2] \\
 & SalFoM (temporal) & 1.090 & [0.900, 1.280] & 58.5 & [52.8, 64.1] \\
 & Gap (SalFoM $-$ DG-MR) & 0.847 & [0.618, 1.077] & 45.5 & [34.7, 56.2] \\
\bottomrule
\end{tabular}
\end{table}

%
%
%

\begin{table}[ht]
\centering
\footnotesize
\caption{\datasetname{} validation results. FT $=$ per-dataset fine-tuning on \datasetname{} train. KLD lower-is-better; LL, IGE, CC, SIM, AUC, NSS higher-is-better. Sorted by IGE descending.}
\label{tab:saltempto-val}
\setlength{\tabcolsep}{2pt}
\begin{tabular}{lcrrrrrrr}
\toprule
Method & FT & LL & IGE & KLD & CC & SIM & AUC & NSS \\
\midrule
Best gold standard (adaptive 4-mix $+$ SalFoM) & --- & 3.1943 & 100.0\,\% & 0.7102 & 0.8175 & 0.5591 & 0.9755 & 8.5661 \\
\midrule
SalFoM                       & \checkmark & 2.3583 & 55.6\,\% & 1.5343 & 0.4816 & 0.3331 & 0.9147 & 3.0683 \\
SalFoM                       &            & 2.2784 & 51.4\,\% & 1.5807 & 0.4702 & 0.3321 & 0.9140 & 2.9986 \\
TMFI-Net                      &            & 2.0917 & 41.5\,\% & 1.8402 & 0.4363 & 0.3091 & 0.9045 & 2.7174 \\
ViNet-E                      &            & 1.9527 & 34.1\,\% & 1.7652 & 0.4016 & 0.2738 & 0.8955 & 2.4230 \\
ViNet-A                      &            & 1.9407 & 33.4\,\% & 1.7607 & 0.4011 & 0.2820 & 0.8955 & 2.4431 \\
UniformerSal-SpatioTemporal  &            & 1.9134 & 32.0\,\% & 1.5251 & 0.4655 & 0.3597 & 0.8983 & 2.4587 \\
TASED-Net                    &            & 1.8595 & 29.1\,\% & 1.8248 & 0.3910 & 0.2705 & 0.8896 & 2.4073 \\
TinyHD-single                &            & 1.8314 & 27.6\,\% & 1.8690 & 0.3884 & 0.2548 & 0.8924 & 2.3710 \\
UniformerSal-Spatial         &            & 1.8210 & 27.1\,\% & 1.6108 & 0.4526 & 0.3562 & 0.8976 & 2.4483 \\
ViNet-S                      &            & 1.7833 & 25.1\,\% & 1.8620 & 0.3719 & 0.2626 & 0.8863 & 2.2508 \\
TinyHD-multi                 &            & 1.7260 & 22.0\,\% & 1.9339 & 0.3671 & 0.2414 & 0.8877 & 2.2273 \\
DG-MR                        & \checkmark & 1.6082 & 15.8\,\% & 2.2170 & 0.3067 & 0.2182 & 0.8719 & 2.0390 \\
UNISAL-Spatial               &            & 1.4617 & \phantom{0}8.0\,\% & 5.9256 & 0.0485 & 0.0088 & 0.8692 & 1.8923 \\
UNISAL-SpatioTemporal        &            & 1.4578 & \phantom{0}7.8\,\% & 2.2547 & 0.3422 & 0.2606 & 0.8757 & 2.0333 \\
DG-MR                        &            & 1.3235 & \phantom{0}0.7\,\% & 2.8716 & 0.2380 & 0.1705 & 0.8467 & 1.9238 \\
\midrule
CenterBias                   & ---        & 1.3110 & \phantom{0}0.0\,\% & 1.8066 & 0.3437 & 0.2527 & 0.8464 & 1.6610 \\
\bottomrule
\end{tabular}
\end{table}

\begin{table}[ht]
\centering
\footnotesize
\caption{\datasetname{} test results, sorted by IGE descending.}
\label{tab:saltempto-test}
\setlength{\tabcolsep}{2pt}
\begin{tabular}{lcrrrrrrr}
\toprule
Method & FT & LL & IGE & KLD & CC & SIM & AUC & NSS \\
\midrule
Best gold standard (adaptive 4-mix $+$ SalFoM) & --- & 3.1579 & 100.0\,\% & 0.7496 & 0.7954 & 0.5523 & 0.9697 & 8.1241 \\
\midrule
SalFoM                       & \checkmark & 2.4462 & 61.4\,\% & 1.4629 & 0.4927 & 0.3500 & 0.9158 & 3.1212 \\
SalFoM                       &            & 2.3653 & 57.0\,\% & 1.5091 & 0.4818 & 0.3433 & 0.9125 & 3.0137 \\
TMFI-Net                      &            & 2.1345 & 44.5\,\% & 1.7996 & 0.4563 & 0.3228 & 0.8995 & 2.8311 \\
TinyHD-single                &            & 1.9555 & 34.8\,\% & 1.7897 & 0.4082 & 0.2665 & 0.8915 & 2.4952 \\
UniformerSal-Spatial         &            & 1.9025 & 31.9\,\% & 1.5533 & 0.4657 & 0.3633 & 0.9002 & 2.4965 \\
ViNet-E                      &            & 1.9025 & 31.9\,\% & 1.7679 & 0.3943 & 0.2798 & 0.8886 & 2.3938 \\
TinyHD-multi                 &            & 1.8904 & 31.3\,\% & 1.8292 & 0.3949 & 0.2604 & 0.8891 & 2.4080 \\
ViNet-S                      &            & 1.8196 & 27.4\,\% & 1.8195 & 0.3744 & 0.2761 & 0.8855 & 2.2970 \\
UniformerSal-SpatioTemporal  &            & 1.7750 & 25.0\,\% & 1.6084 & 0.4527 & 0.3531 & 0.8975 & 2.3957 \\
TASED-Net                    &            & 1.7607 & 24.2\,\% & 1.8604 & 0.3810 & 0.2691 & 0.8805 & 2.3187 \\
ViNet-A                      &            & 1.7435 & 23.3\,\% & 1.8420 & 0.3774 & 0.2758 & 0.8797 & 2.2815 \\
UNISAL-Spatial               &            & 1.5029 & 10.3\,\% & 5.9185 & 0.0496 & 0.0087 & 0.8706 & 1.9676 \\
DG-MR                        & \checkmark & 1.5016 & 10.2\,\% & 2.2650 & 0.2860 & 0.2076 & 0.8547 & 1.8680 \\
UNISAL-SpatioTemporal        &            & 1.4126 & \phantom{0}5.4\,\% & 2.2780 & 0.3406 & 0.2598 & 0.8656 & 2.0571 \\
\midrule
CenterBias                   & ---        & 1.3136 & \phantom{0}0.0\,\% & 1.8003 & 0.3268 & 0.2431 & 0.8479 & 1.6369 \\
DG-MR                        &            & 1.2052 & $-5.9$\,\% & 2.4810 & 0.2538 & 0.2067 & 0.8336 & 1.6875 \\
\bottomrule
\end{tabular}
\end{table}

\begin{table}[ht]
\centering
\footnotesize
\caption{LEDOV validation results. IGE is the fraction of the centerbias-to-gold-standard headroom recovered. FT $=$ per-dataset fine-tuning on LEDOV train; $\dagger$ released checkpoint already trained on LEDOV by the original authors (no further fine-tuning in this work). KLD lower-is-better; LL, IGE, CC, SIM, AUC, NSS higher-is-better. Sorted by IGE descending. Gold-standard centerbias and gold LL: 1.2560 / 3.6948.}
\label{tab:ledov-val}
\setlength{\tabcolsep}{2pt}
\begin{tabular}{lcrrrrrrr}
\toprule
Method & FT & LL & IGE & KLD & CC & SIM & AUC & NSS \\
\midrule
Best gold standard (adaptive 4-mix $+$ SalFoM) & ---                  & 3.6948 & 100.0\,\% & 0.4009 & 0.8861 & 0.7117 & 0.9808 & 9.6380 \\
\midrule
SalFoM                       & \checkmark                              & 3.0937 & 75.4\,\% & 0.9234 & 0.7289 & 0.5401 & 0.9368 & 4.5045 \\
UniformerSal-SpatioTemporal  & \checkmark$^\dagger$                    & 2.8710 & 66.2\,\% & 1.0990 & 0.6630 & 0.4900 & 0.9300 & 4.2670 \\
SalFoM                       &                                         & 2.8472 & 65.3\,\% & 1.0556 & 0.6919 & 0.5031 & 0.9254 & 4.2242 \\
TMFI-Net                      &                                         & 2.7959 & 63.1\,\% & 1.0804 & 0.6855 & 0.4732 & 0.9244 & 4.0483 \\
UniformerSal-Spatial         & \checkmark$^\dagger$                    & 2.6471 & 57.0\,\% & 1.0030 & 0.6778 & 0.5353 & 0.9203 & 3.9882 \\
DG-MR                        & \checkmark$^\dagger$                    & 2.6115 & 55.6\,\% & 1.2254 & 0.6425 & 0.4673 & 0.9174 & 3.8552 \\
TinyHD-single                &                                         & 2.5938 & 54.9\,\% & 1.2062 & 0.6580 & 0.4320 & 0.9153 & 3.9068 \\
TinyHD-multi                 &                                         & 2.5237 & 52.0\,\% & 1.2511 & 0.6371 & 0.4148 & 0.9133 & 3.7798 \\
ViNet-E                      &                                         & 2.2565 & 41.0\,\% & 1.3712 & 0.5770 & 0.4137 & 0.9017 & 3.2337 \\
ViNet-S                      &                                         & 2.2284 & 39.9\,\% & 1.4052 & 0.5644 & 0.4159 & 0.9007 & 3.2266 \\
ViNet-A                      &                                         & 2.1614 & 37.1\,\% & 1.4385 & 0.5645 & 0.4087 & 0.8968 & 3.2265 \\
TASED-Net                    &                                         & 2.0454 & 32.4\,\% & 1.4792 & 0.5428 & 0.3575 & 0.8909 & 2.8951 \\
UNISAL-Spatial               &                                         & 1.9983 & 30.4\,\% & 1.5628 & 0.5338 & 0.3100 & 0.8958 & 2.9333 \\
UNISAL-SpatioTemporal        &                                         & 1.9836 & 29.8\,\% & 1.6098 & 0.5470 & 0.4107 & 0.8908 & 3.1071 \\
\midrule
CenterBias                   & ---                                     & 1.2560 & \phantom{0}0.0\,\% & 1.7238 & 0.3554 & 0.2925 & 0.8416 & 1.6506 \\
\bottomrule
\end{tabular}
\end{table}

\begin{table}[ht]
\centering
\footnotesize
\caption{LEDOV test results. IGE is the fraction of the centerbias-to-gold-standard headroom recovered. FT $=$ per-dataset fine-tuning on LEDOV train; $\dagger$ released checkpoint already trained on LEDOV by the original authors (no further fine-tuning in this work). KLD lower-is-better; LL, IGE, CC, SIM, AUC, NSS higher-is-better. Sorted by IGE descending. Gold-standard centerbias and gold LL: 1.3329 / 3.6209.}
\label{tab:ledov-test}
\setlength{\tabcolsep}{2pt}
\begin{tabular}{lcrrrrrrr}
\toprule
Method & FT & LL & IGE & KLD & CC & SIM & AUC & NSS \\
\midrule
Best gold standard (adaptive 4-mix $+$ SalFoM) & ---                  & 3.6209 & 100.0\,\% & 0.3955 & 0.8825 & 0.7122 & 0.9803 & 9.5950 \\
\midrule
SalFoM                       & \checkmark                              & 3.0524 & 75.2\,\% & 0.8969 & 0.7341 & 0.5372 & 0.9378 & 4.3478 \\
UniformerSal-SpatioTemporal  & \checkmark$^\dagger$                    & 2.9056 & 68.7\,\% & 0.8218 & 0.7361 & 0.5766 & 0.9329 & 4.2549 \\
SalFoM                       &                                         & 2.8309 & 65.5\,\% & 1.0163 & 0.6920 & 0.5139 & 0.9288 & 4.0333 \\
TMFI-Net                      &                                         & 2.7860 & 63.5\,\% & 1.0408 & 0.6864 & 0.4780 & 0.9280 & 3.8610 \\
UniformerSal-Spatial         & \checkmark$^\dagger$                    & 2.7021 & 59.8\,\% & 0.9206 & 0.6886 & 0.5443 & 0.9258 & 3.8798 \\
DG-MR                        & \checkmark$^\dagger$                    & 2.6241 & 56.4\,\% & 1.1536 & 0.6427 & 0.4748 & 0.9201 & 3.6546 \\
TinyHD-single                &                                         & 2.5802 & 54.5\,\% & 1.1690 & 0.6472 & 0.4350 & 0.9202 & 3.6665 \\
TinyHD-multi                 &                                         & 2.5678 & 54.0\,\% & 1.1756 & 0.6418 & 0.4284 & 0.9215 & 3.6138 \\
ViNet-E                      &                                         & 2.3962 & 46.5\,\% & 1.2477 & 0.5859 & 0.4298 & 0.9152 & 3.1637 \\
ViNet-S                      &                                         & 2.3582 & 44.8\,\% & 1.2868 & 0.5727 & 0.4320 & 0.9135 & 3.1447 \\
ViNet-A                      &                                         & 2.3095 & 42.7\,\% & 1.3134 & 0.5646 & 0.4198 & 0.9113 & 3.0966 \\
TASED-Net                    &                                         & 2.1679 & 36.5\,\% & 1.3634 & 0.5636 & 0.3760 & 0.9046 & 2.9020 \\
UNISAL-SpatioTemporal        &                                         & 2.0565 & 31.6\,\% & 1.5136 & 0.5431 & 0.4192 & 0.8998 & 2.9770 \\
UNISAL-Spatial               &                                         & 1.9864 & 28.6\,\% & 1.5255 & 0.5315 & 0.3136 & 0.9011 & 2.8260 \\
\midrule
CenterBias                   & ---                                     & 1.3329 & \phantom{0}0.0\,\% & 1.6527 & 0.3683 & 0.3011 & 0.8515 & 1.6875 \\
\bottomrule
\end{tabular}
\end{table}

\xhdr{Cross-dataset rank consistency}
Spearman $\rho$ between LEDOV and \datasetname{} IG rankings (each model's val and test IG pooled to a single score per dataset), computed over the full \Nmodels{}-model panel, is $\rho = \lockedRho{}$. Excluding the three models whose released checkpoints were trained on LEDOV by their original authors (DG-MR, UniformerSal-Spatial, UniformerSal-SpatioTemporal) raises Spearman $\rho$ to \lockedRhoExcluded{} on the 10-model zero-shot inference subset (used in Figure~\ref{fig:cross_dataset_zeroshot}).

\section{Control experiments}
\label{app:gs-substitution}


\xhdr{Substituting the deep component of the gold standard}
Table~\ref{tab:gs_substitution} replaces the fine-tuned SalFoM component of the gold-standard mixture (\S\ref{sec:methods:goldstandard}) with each of the next-strongest panel models and recomputes the ceiling and the recoveries of the two fine-tuned models. The ceiling moves by at most $0.11$\,bit, and the static-vs-temporal gap by at most $1.4$ points on either dataset, so the cross-dataset contrast does not depend on which model serves as the component. SalFoM recovers the smallest fraction of the headroom under the SalFoM-based ceiling on both datasets, because a stronger component raises the ceiling that every model, itself included, is measured against; the reported configuration is therefore the most demanding of the four. Note that the key metrics based on fixation coordinates (LL and IG) do not depend on the gold-standard mixture and are are directly computed based on model outputs, while IGE contains the model likelihoods as a normalizing factor across the whole panel being evaluated; there is no circularity in including the model in the gold-standard mixture.

\begin{table}[ht]
\centering
\small
\caption{Gold-standard sensitivity to the deep mixture component. Ceiling is the gold-standard IG over the centerbias in bits, pooled over validation and test videos; SalFoM and DG-MR columns give the IGE of the fine-tuned models under that ceiling, and gap is their difference in percentage points.}
\label{tab:gs_substitution}
\setlength{\tabcolsep}{4pt}
\begin{tabular}{lrrrrrrrr}
\toprule
 & \multicolumn{4}{c}{LEDOV} & \multicolumn{4}{c}{\datasetname{}} \\
\cmidrule(lr){2-5} \cmidrule(lr){6-9}
Deep component & ceiling & SalFoM & DG-MR & gap & ceiling & SalFoM & DG-MR & gap \\
\midrule
SalFoM (reported)    & 2.359 & 75.2\,\% & 56.0\,\% & 19.2 & 1.864 & 58.5\,\% & 13.0\,\% & 45.5 \\
UniformerSal         & 2.328 & 76.2\,\% & 56.7\,\% & 19.5 & 1.831 & 59.5\,\% & 13.3\,\% & 46.3 \\
DeepGaze MR          & 2.283 & 77.7\,\% & 57.9\,\% & 19.9 & 1.808 & 60.3\,\% & 13.4\,\% & 46.8 \\
TMFI-Net              & 2.252 & 78.8\,\% & 58.7\,\% & 20.2 & 1.848 & 59.0\,\% & 13.1\,\% & 45.8 \\
\bottomrule
\end{tabular}
\end{table}

\xhdr{SalFoM without temporal context}
\label{app:static-salfom}
To separate SalFoM's static features from its use of temporal context, we evaluate the fine-tuned model with the target frame repeated across its full context window, so that every position in the window shows the same image and the input carries no motion (Table~\ref{tab:static_salfom}). On LEDOV the static condition still recovers 64\,\% of the headroom, and removing temporal context costs SalFoM 0.26\,bit, 15\,\% of its information gain over the centerbias. On \datasetname{} the static condition recovers 33\,\%, and removing temporal context costs 0.48\,bit, 44\,\% of its gain. The share of SalFoM's margin over DG-MR that is attributable to temporal context is similar on both datasets (57\,\%), because DG-MR is also weaker on \datasetname{}; the contrast between the datasets lies in how much of SalFoM's own gain depends on temporal context and in how much a static model with SalFoM's backbone can recover. Because a repeated frame is an input the model never saw in training, the static condition is a lower bound on what such a static model could achieve. The LEDOV drop bounds this artefact from above: if the entire LEDOV drop were artefact and it were of the same size on \datasetname{}, at least 0.22\,bit of the \datasetname{} drop would remain attributable to temporal information.
\begin{table}[ht]
\centering
\small
\caption{SalFoM with and without temporal context. The frame-repeated row feeds the fine-tuned SalFoM the target frame at every position of its context window. IG is over the centerbias in bits per fixation, pooled over the released evaluation videos of each dataset (77 on LEDOV, 20 on \datasetname{}); the IGE column gives the share of the centerbias-to-gold headroom (Eq.~\ref{eq:ige}), or IGE points for the two difference rows. The three model rows within a dataset are scored on the same videos and carry 95\,\% intervals for paired comparisons among them, obtained by removing each video's mean over the three rows \citep{cousineau_confidence_2005,morey_confidence_2008}; the difference rows are paired over the same videos with 95\,\% $t$-intervals. Intervals for absolute IG are in Table~\ref{tab:jackknife}.}
\label{tab:static_salfom}
\begin{tabular}{llrlr}
\toprule
Dataset & Row & IG (bits) & 95\,\% interval & IGE (\%) \\
\midrule
LEDOV & DG-MR (fine-tuned) & 1.321 & [1.254, 1.389] & 56.0 \\
 & SalFoM, frame repeated & 1.515 & [1.452, 1.579] & 64.3 \\
 & SalFoM (fine-tuned) & 1.775 & [1.725, 1.824] & 75.2 \\
 & Temporal context (SalFoM $-$ frame repeated) & 0.259 & [0.185, 0.333] & 11.0 \\
 & Static features (frame repeated $-$ DG-MR) & 0.194 & [0.095, 0.293] & 8.2 \\
\midrule
\datasetname{} & DG-MR (fine-tuned) & 0.243 & [0.072, 0.414] & 13.0 \\
 & SalFoM, frame repeated & 0.609 & [0.506, 0.712] & 32.7 \\
 & SalFoM (fine-tuned) & 1.090 & [0.965, 1.215] & 58.5 \\
 & Temporal context (SalFoM $-$ frame repeated) & 0.481 & [0.357, 0.605] & 25.8 \\
 & Static features (frame repeated $-$ DG-MR) & 0.366 & [0.159, 0.573] & 19.7 \\
\bottomrule
\end{tabular}
\end{table}

\xhdr{DIEM}
\label{app:diem}
On DIEM \citep{mital_clustering_2011}, evaluated zero-shot with the same pipeline on its 17 test videos, DG-MR recovers $\approx$22\,\% of the gold-standard headroom and SalFoM $\approx$54\,\%. DIEM's gaze was recorded with audio playback and heterogeneous equipment, and its model ranking correlates only weakly with the other two datasets (Spearman $\rho = 0.44$ with LEDOV and $0.62$ with \datasetname{}), so we treat it as a secondary check.

\section{Dataset-level dynamism analysis}
\label{app:dynamism}

The clearest gaze-anchored signature of within-video dispersion is the cross-frame transfer log-likelihood (LL): one frame's gold-standard density evaluated on another frame's fixations, averaged across within-video frame pairs (Table~\ref{tab:dynamism}). On LEDOV, cross-frame transfer carries roughly the same predictive information as the dataset-wide centerbias: LL is $+1.06$ vs CB $+1.26$ bits per fixation on validation ($+1.21$ vs $+1.33$ on test). This indicates that within-video gaze targets are clustered tightly enough that a frame-agnostic predictor is a reasonable estimate of where salient regions in other frames lie. On \datasetname{}, the same quantity drops below centerbias by over a bit: LL $-0.06$ vs CB $+1.31$ on validation ($+0.09$ vs $+1.31$ on test). Substituting one \datasetname{} frame's gold density for another's costs $\approx$1.4 bits per fixation relative to centerbias: the marker of highly dynamic within-video gaze targets.

\xhdr{Cross-frame recovery as a gold-relative fraction}
\label{app:dynamism:crossframerecov}
The cross-frame transfer LL can be recast as a fraction of the achievable per-frame predictability (gold-standard LL above the uniform spatial baseline). On LEDOV val, gold LL above uniform is $\approx$3.77 bits per fixation, so one frame's gold density evaluated on another within-video frame recovers $1.06 / 3.77 \approx 28\%$ of the achievable per-frame log-likelihood (\S\ref{sec:intro}). On \datasetname{} val, cross-frame transfer falls below the uniform baseline (LL $-0.06$), so the same fraction is non-positive: substituting one frame's gold density for another's actively misleads, rather than transferring usefully, necessitating models to actually track salient regions over time to make accurate predictions.

The pairwise symmetrised KL between per-frame gold-standard densities tells the same story from a different reference frame: $\approx$30\,\% larger on \datasetname{} ($3.11$ vs $2.38$ bits validation; $2.98$ vs $2.17$ test). RAFT optical-flow magnitude \citep{teed_raft_2020} at matched evaluation resolution (long edge 512) is $\approx$2.5$\times$ larger on \datasetname{} (mean $2.53$ vs $1.00$ px/frame validation), further confirming high spatial dynamism, although pixel-level metrics on HACS-derived web video are partially confounded by camera shake. The dispersion of optical flow across videos is also informative: its coefficient of variation is $\approx$1.6 on LEDOV but $\approx$0.56 on \datasetname{}, so LEDOV's dynamism is concentrated in a few highly dynamic videos while \datasetname{}'s is more uniform across the dataset, which indicates uniform dynamism rather than a few highly dynamic outlier frames. All three metrics agree on direction; the gaze-anchored cross-frame LL is the most directly diagnostic.

\begin{table}[ht]
\centering
\small
\caption{Dataset-level dynamism. KL is the pairwise symmetrised divergence (in bits) between per-frame gold-standard densities, averaged over within-video frame pairs (diagonal excluded) and across videos. LL is the cross-frame transfer log-likelihood: one frame's gold-standard density evaluated on another frame's fixations, in bits above a uniform spatial baseline, averaged over within-video frame pairs and across videos. RAFT optical-flow magnitudes are at matched evaluation resolution (long edge 512).}
\label{tab:dynamism}
\begin{tabular}{l c c c}
\toprule
Split & KL (bits) & LL (bits) & Optical flow (px/frame) \\
\midrule
LEDOV val   & $2.378 \pm 1.059$ & $\phantom{-}1.057 \pm 0.887$ & $0.995 \pm 1.619$ \\
LEDOV test  & $2.170 \pm 0.997$ & $\phantom{-}1.212 \pm 0.808$ & $1.213 \pm 2.131$ \\
\datasetname{} val  & $3.106 \pm 1.059$ & $-0.055 \pm 0.499$            & $2.529 \pm 1.413$ \\
\datasetname{} test & $2.979 \pm 0.804$ & $\phantom{-}0.093 \pm 0.611$ & $3.080 \pm 2.226$ \\
\bottomrule
\end{tabular}
\end{table}

\end{document}